\pdfoutput=1

\documentclass[11pt]{article}

\usepackage[final]{acl}

\usepackage{times}
\usepackage{latexsym}
\usepackage{booktabs}
\usepackage[dvipsnames,table]{xcolor}
\usepackage[T1]{fontenc}
\usepackage[most]{tcolorbox}
\usepackage{booktabs}
\usepackage{graphicx}
\usepackage{caption}
\usepackage{adjustbox}
\usepackage{multirow}
\usepackage{colortbl}
\usepackage{lscape}
\usepackage{array}
\usepackage{courier}
\usepackage{enumitem}
\usepackage{pifont}
\usepackage{placeins}
\usepackage{inconsolata}
\usepackage{graphicx}
\usepackage{times}
\usepackage{latexsym}
\usepackage{algorithm}
\usepackage{algorithmic}
\usepackage{amsmath}
\usepackage{ragged2e}
\usepackage{xcolor}
\usepackage[most]{tcolorbox}
\usepackage{rotating}
\usepackage{tablefootnote}
\usepackage{amssymb}
\usepackage{textcomp}

\definecolor{BlueThirty}{RGB}{77,77,255}
\definecolor{Grey}{gray}{0.9} 

\newcommand\datasetname{\textcolor{black}{\textsc{PingPong}}}

\newcommand{\colorIndicator}[1]{%
  \raisebox{0.5ex}{%
    \tcbox[colback=#1,
           colframe=#1,
           arc=0pt,
           boxrule=0pt,
           left=2pt,
           right=2pt,
           top=0pt,
           bottom=0pt,
           width=3mm,
           height=2ex,
           valign=center,
           on line]{}%
  }%
  \space%
}

\newcommand{\tagbox}[2]{%
  \tcbox[colback=#1!20,
         colframe=#1,
         arc=5pt,
         boxrule=0.8pt,
         left=2pt,
         right=2pt,
         top=1pt,
         bottom=1pt,
         on line]{\small{#2}}%
}

\newcommand{\cmark}{\tcbox[colback=Green!50,colframe=Green,arc=4pt,boxrule=0pt,left=0.5pt,right=0.5pt,top=0.5pt,bottom=0.5pt, on line]{\ding{51}}} 
\newcommand{\xmark}{\tcbox[colback=Grey!20,colframe=Red,arc=4pt,boxrule=0pt,left=0.5pt,right=0.5pt,top=0.5pt,bottom=0.5pt, on line]{\ding{55}}} 

\usepackage[utf8]{inputenc}

\usepackage{microtype}

\usepackage{inconsolata}

\usepackage{graphicx}

\title{\datasetname: A Natural Benchmark for Multi-Turn Code-Switching Dialogues}

\usepackage{skak}

\def\quad{\hskip0.75em\relax}

\author{
    \textbf{Mohammad Rifqi Farhansyah}$^{1,2*}$, \textbf{Hanif Muhammad Zhafran}$^{1*}$, \textbf{Farid Adilazuarda}$^{3}$,\\
    \textbf{Shamsuddeen Hassan Muhammad}$^{6}$,
    \textbf{Maryam Ibrahim Mukhtar}$^{8}$, \textbf{Nedjma Ousidhoum}$^{7}$, \\
    \textbf{Sathwik Reddy}$^{10}$, \textbf{Ponnurangam Kumaraguru}$^{10}$,
    \textbf{Prashant Kodali}$^{9}$, \textbf{Sunayana Sitaram}$^{9}$, \\
    \textbf{Genta Indra Winata}$^{5\dagger}$, \textbf{Ayu Purwarianti}$^{1\dagger}$, \textbf{Alham Fikri Aji}$^{4\dagger}$ \\
    $^1$Bandung Institute of Technology \quad
    $^2$Monash University Indonesia \quad
    $^3$University of Edinburgh \\
    $^4$MBZUAI \quad
    $^5$Capital One \quad
    $^6$Imperial College London \quad 
    $^7$Cardiff University \\
    $^8$Bayero University Kano \quad
    $^9$Microsoft Research, India \quad
    $^{10}$IIIT Hyderabad
    \\
    \texttt{\{mrifqifarhansyah, hanif.zhafran07\}@gmail.com}
    \\
    \normalsize{$^{*}$Main Author}\quad
    \normalsize{$^{\dagger}$Senior Author}\quad
}
\vspace{1.5em}

\begin{document}
\maketitle
\begin{abstract}
Code-switching is a widespread practice among the world’s multilingual majority, yet few benchmarks accurately reflect its complexity in everyday communication. We present $\datasetname$, a benchmark for natural multi-party code-switching dialogues covering five language-combination variations, some of which are trilingual. Our dataset consists of human-authored conversations among 2 to 4 participants covering authentic, multi-threaded structures where replies frequently reference much earlier points in the dialogue. We demonstrate that our data is significantly more natural and structurally diverse than machine-generated alternatives, offering greater variation in message length, speaker dominance, and reply distance. Based on these dialogues, we define three downstream tasks: \textsc{Question Answering}, \textsc{Dialogue Summarization}, and \textsc{Topic Classification}. Evaluations of several state-of-the-art language models on $\datasetname$ reveal that performance remains limited on code-switched inputs, underscoring the urgent need for more robust NLP systems capable of addressing the intricacies of real-world multilingual discourse.
\end{abstract}

\section{Introduction}


\begin{figure}[!th]
  \centering
  \setlength{\abovecaptionskip}{3pt}
  \setlength{\abovecaptionskip}{3pt}
  \includegraphics[width=0.4\textwidth]{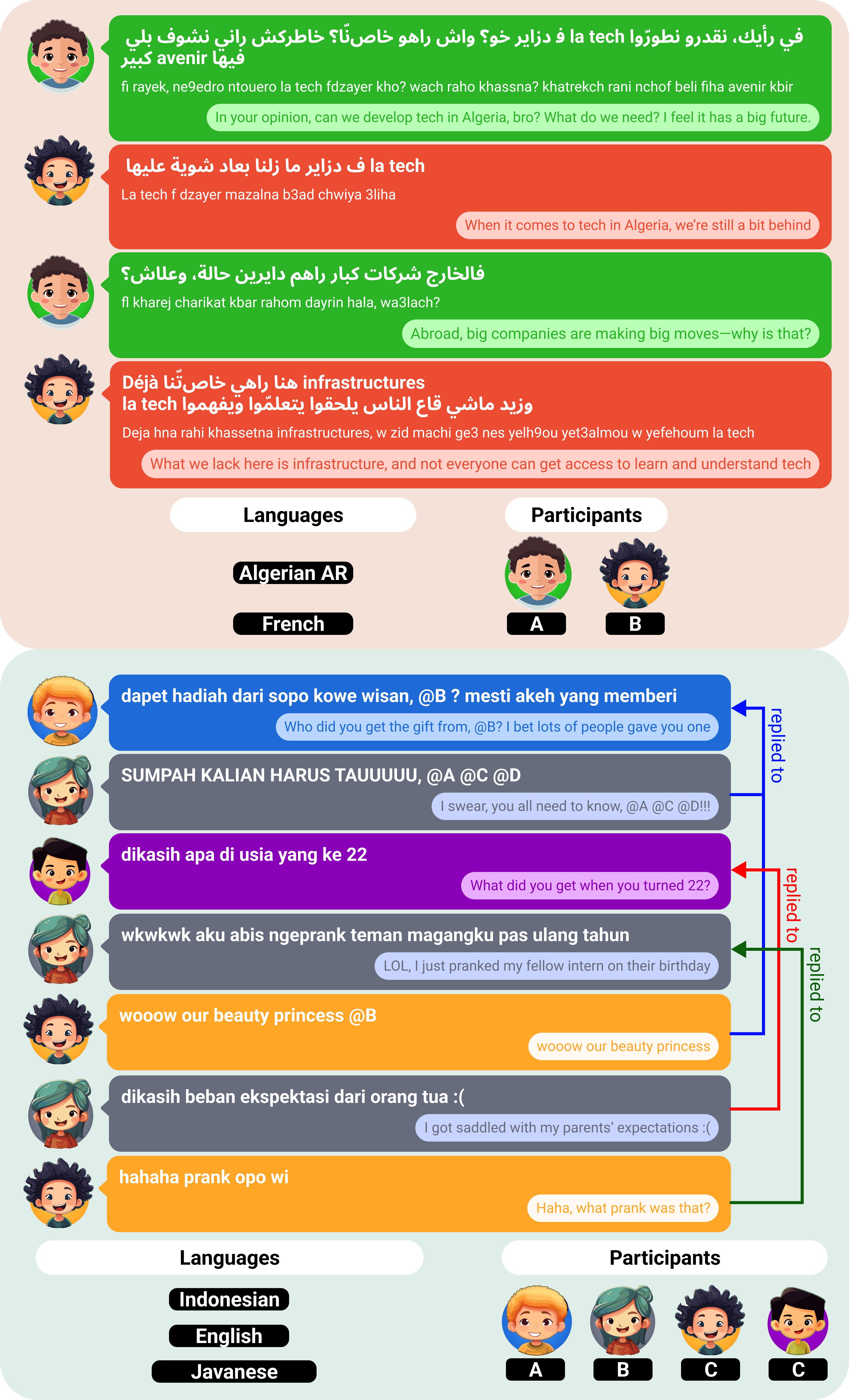}
  \caption{Illustration of dialogue samples from $\datasetname$. The bottom panel highlights how each turn corresponds to previous turns, as labeled in our dataset.}
  \label{fig:dataset_coverage}
  \vspace{-1em}
\end{figure}

\begin{table*}[!th]
\centering
\setlength{\abovecaptionskip}{3pt}
\renewcommand{\arraystretch}{1.25}
\resizebox{0.95\textwidth}{!}{
\begin{tabular}{lcccccc}
\toprule
\textbf{Dataset} & \textbf{Open-Source} & \textbf{CS Type} & \textbf{\#Comp.\ Lang.} & \textbf{\#Regions} & \textbf{\#Gen. Task} & \textbf{Conv.} \\

\midrule

SEAME ~\cite{lyu2010seame} & \cmark & \tagbox{Grey}{Bilingual} & 0 & 1 & 0 & \cmark \\

GLUECoS ~\cite{khanuja-etal-2020-gluecos} & \cmark & \tagbox{Grey}{Bilingual} & 0 & 2 & 1 &  \xmark \\




CommonDost~\cite{parekh-etal-2020-understanding} & \cmark & \tagbox{Grey}{Bilingual} & 0 & 2 & 1 & \cmark\\




MalayalamMixSentiment~\cite{chakravarthi-etal-2020-sentiment} & \cmark & \tagbox{Grey}{Bilingual} & 1 & 1 & 0 & \xmark \\




TamilMixSentiment~\cite{chakravarthi-etal-2020-corpus} & \xmark & \tagbox{Grey}{Bilingual} & 0 & 1 & 0 & \xmark \\




hinglishNorm~\cite{makhija-etal-2020-hinglishnorm} & \cmark & \tagbox{Grey}{Bilingual} & 0 & 1 & 1 & \cmark\\




LinCE ~\cite{aguilar-etal-2020-lince} & \cmark & \tagbox{Grey}{Bilingual} & 2 & 3 & 1 & \xmark \\




CSCS~\cite{balabel-etal-2020-cairo} & \cmark & \tagbox{Grey}{Bilingual} & 1 & 1 & 0 & \xmark \\




CanVEC~\cite{nguyen-bryant-2020-canvec} & \cmark & \tagbox{Grey}{Bilingual} & 1 & 1 & 1 & \xmark \\




ArzEn~\cite{hamed-etal-2020-arzen} & \xmark & \tagbox{Grey}{Bilingual} & 1 & 1 & 1 & \xmark \\




PHINC~\cite{srivastava-singh-2020-phinc} & \cmark & \tagbox{Grey}{Bilingual} & 0 & 1 & 1 & \xmark \\




MIPE~\cite{garg-etal-2021-mipe} & \xmark & \tagbox{Grey}{Bilingual} & 0 & 1 & 1 & \xmark \\




HinGE~\cite{srivastava-singh-2021-hinge} & \cmark & \tagbox{Grey}{Bilingual} & 0 & 1 & 1 & \xmark \\




TCS~\cite{tarunesh-etal-2021-machine} & \xmark & \tagbox{Grey}{Bilingual} & 0 & 1 & 2 & \cmark \\




GupShup~\cite{mehnaz-etal-2021-gupshup} & \cmark & \tagbox{Grey}{Bilingual} & 0 & 1 & 1 & \cmark \\




DOSA~\cite{ravikiran-annamalai-2021-dosa} & \cmark & \tagbox{Grey}{Bilingual} & 1 & 1 & 0 & \xmark \\




TweetTaglish~\cite{herrera-etal-2022-tweettaglish} & \cmark & \tagbox{Grey}{Bilingual} & 1 & 1 & 0 & \xmark\\




BaSCo~\cite{aguirre-etal-2022-basco} & \cmark & \tagbox{Grey}{Bilingual} & 1 & 1 & 0 & \cmark\\




MHE~\cite{rani-etal-2022-mhe} & \xmark & \tagbox{Green}{Trilingual} & 1 & 1 & 0 & \xmark \\




ASCEND~\cite{lovenia2022ascend} & \cmark & \tagbox{Grey}{Bilingual} & 0 & 1 & 1 & \cmark \\




L3Cube-HingCorpus~\cite{nayak-joshi-2022-l3cube} & \cmark & \tagbox{Grey}{Bilingual} & 0 & 1 & 1 & \xmark \\




CroCoSum~\cite{zhang-eickhoff-2024-crocosum} & \cmark & \tagbox{Grey}{Bilingual} & 0 & 1 & 1 & \xmark \\

CS-PREF~\cite{kuwanto2024linguistics} & \cmark & \tagbox{Grey}{Bilingual} & 1 & 2 & 1 & \xmark \\


CS-Sum~\cite{suresh2025cs} & \cmark & \tagbox{Grey}{Bilingual} & 1 & 2 & 1 & \xmark \\


CodeMixQA~\cite{winata2026can} & \cmark & \tagbox{Grey}{Bilingual} & 16 & 3 & 1 & \xmark \\

\midrule
\textbf{\datasetname} & \cmark & \tagbox{Grey}{Bilingual} \tagbox{Green}{Trilingual} & \textbf{4} & \textbf{3} & \textbf{2} &  \textbf{\cmark} \\


\bottomrule
\end{tabular}
}
\caption{Comparison of code-switching benchmark datasets. \textbf{Open-Source} indicates public availability of the dataset. \textbf{CS Type} denotes the code-switching setting. \textbf{\# Comp. Lang.} counts under-studied languages beyond high-resource ones (e.g., English, Hindi, Chinese). \textbf{\# Regions} reports geographic coverage (e.g., South Asian-English, European-English). \textbf{\# Gen. Tasks} indicates the number of generative tasks. \textbf{Conv.} denotes whether the dataset is conversational.}
\label{tab:comparison_benchmark}
\end{table*}

Code-switching, the practice of alternating between two or more languages within a single conversation, is a pervasive linguistic phenomenon in contemporary multilingual societies~\cite{poplack2013sometimes, myers1997duelling, bullock2009cambridge, auer2013code}. With the global number of multilingual speakers surpassing that of monolinguals, the frequency of code-switching in both informal and formal communication continues to rise~\cite{tucker1999global, grosjean2010bilingual, wei2018translanguaging, winata2021multilingual}. Despite recent progress in multilingual language models~\cite{xue2020mt5, aryabumi2024aya, team2025gemma, yang2025qwen3}, their ability to handle code-switched dialogue remains underexplored and insufficiently evaluated~\cite{li2012code,winata2018code, adilazuarda-etal-2022-indorobusta}. Although several code-switching benchmarks have been introduced in the past, these resources are becoming outdated, and therefore, increasingly misaligned with the current era of LLMs and are less reflective of the everyday realities of code-switching~\cite{aguilar2020lince, khanuja2020gluecos}. 

To address this gap, we present $\datasetname$, a novel benchmark for code-switching in dialogues. Our benchmark consists of conversations in which fluent multilingual speakers naturally code-switch between languages while discussing predefined topics, resulting in multiple language-combination variations through multi-party conversational interactions. $\datasetname$ comprises three downstream tasks, including \textsc{Question Answering}, \textsc{Dialogue Summarization}, and \textsc{Topic Classification}. In addition, the benchmark provides broad language coverage, spanning both high-resource and low-resource language combinations, with certain language combination further distinguished by differences in writing scripts.

In summary, our contributions are as follow:

\begin{figure*}[!t]
  \centering
  \setlength{\abovecaptionskip}{3pt}
  \includegraphics[width=0.9\textwidth]{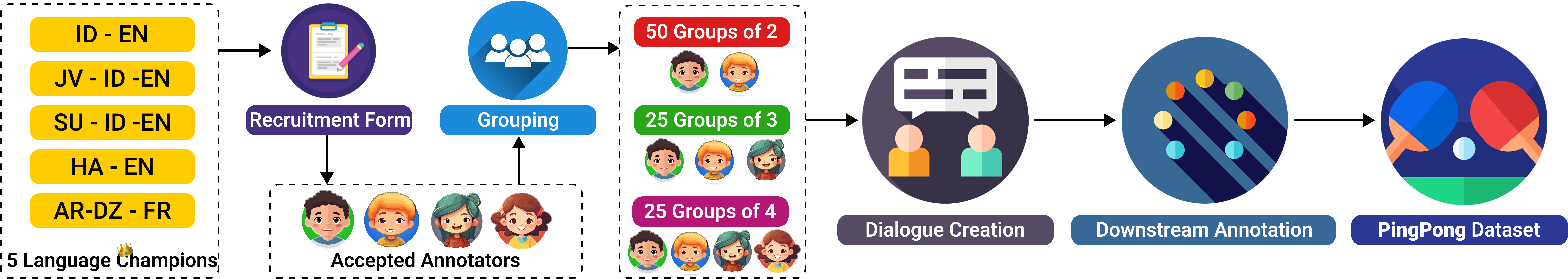}
  \caption{Overview of $\datasetname$ dataset construction. We recruit annotators who speak five language combinations and group them into small teams of 2 to 4 participants. Each group engages in multi-party dialogue creation, after which downstream annotations are performed. The final output is the curated $\datasetname$ dataset, focusing on natural code-switched dialogues for multiple tasks.}
  \label{fig:dataset_construction}
\end{figure*}

\begin{itemize}
    \item We introduce $\datasetname$~\footnote{https://huggingface.co/code-switching}, a novel benchmark for evaluating natural multi-party code-switching in conversational settings, designed to capture the fluid dynamics of multilingual communication.
    \item $\datasetname$ consists of authentic interactions among 2 to 4 speakers discussing everyday topics, with conversations ranging from brief exchanges of 17 turns to extended dialogues of up to 189 turns. Reflecting real conversational behavior, speakers may produce long-distance replies by responding to earlier messages while alternating turns.
    \item $\datasetname$ covers five language combinations spanning high-resource, low-resource, and diverse-script languages, enabling evaluation across a broad range of multilingual scenarios.
    \item Our evaluation framework spans three downstream tasks—\textsc{Question Answering}, \textsc{Dialogue Summarization}, and \textsc{Topic Classification}—providing a holistic assessment of language understanding and generation under code-switching conditions.
    \item We conduct extensive experiments with both open-source and proprietary LLMs, revealing their limitations and highlighting persistent challenges in modeling code-switched, multi-party conversations.
\end{itemize}

\section{Related Work}

Research on code-switching has been studied extensively by linguists for decades~\cite{sitaram2019survey}. In NLP, code-switching has been explored across a wide range of tasks~\cite{winata2023decades}, including language identification~\cite{aguilar2020english, burchell2024code, ojo2025divers, das2023improving}, named entity recognition~\cite{winata2019learning, aguilar2018named, jain2018simple}, part-of-speech tagging~\cite{soto2018joint, ccetinouglu2016part, ball2018part}, and sentiment analysis~\cite{zhang2021cross, shynkarov2025improving}. In parallel, dataset development and shared evaluations have been actively promoted through initiatives such as CALCS~\cite{heredia2025euska, babatunde2025beyond} and the FIRE shared tasks~\cite{kulkarni2024l3cube, adeyemi2023ciral, mandl2024overview}. More recently, this line of work has begun to consolidate around shared benchmarks that enable systematic comparison across models, language combinations, and tasks.

However, despite the growing body of work, several limitations remain across existing code-switching datasets (Table~\ref{tab:comparison_benchmark}). From a sustainability perspective, a number of datasets are not publicly released~\cite{chakravarthi-etal-2020-corpus, hamed-etal-2020-arzen, garg-etal-2021-mipe}. In terms of linguistic diversity, most benchmarks focus exclusively on bilingual code-switching, with only a single dataset explicitly addressing trilingual scenarios~\cite{rani-etal-2022-mhe}. Moreover, both the language and regional coverage remain narrow, as code-switching phenomena in many languages are still under-explored in NLP~\cite{chakravarthi-etal-2020-sentiment, aguilar-etal-2020-lince, balabel-etal-2020-cairo, nguyen-bryant-2020-canvec, hamed-etal-2020-arzen, ravikiran-annamalai-2021-dosa, herrera-etal-2022-tweettaglish, aguirre-etal-2022-basco, rani-etal-2022-mhe, kuwanto2024linguistics, suresh2025cs, winata2026can}.

In addition, most existing datasets fall short of capturing the conversational nature of real-world code-switching, which often emerges in spontaneous and interactive settings~\cite{lyu2010seame, parekh-etal-2020-understanding, makhija-etal-2020-hinglishnorm, tarunesh-etal-2021-machine, mehnaz-etal-2021-gupshup, lovenia2022ascend}. Evaluation tasks are also frequently limited to traditional benchmarks that no longer reflect contemporary NLP challenges~\cite{khanuja-etal-2020-gluecos, nayak-joshi-2022-l3cube}. To address these gaps, we introduce $\datasetname$, a large-scale open-source evaluation suite that expands language and regional coverage, incorporates diverse and realistic conversational code-switching scenarios, and emphasizes generative and semantic tasks.

\section{$\datasetname$ Dataset}

Our dataset is constructed through manual crowdsourcing (Figure~\ref{fig:dataset_construction}), with dialogues collected and downstream tasks curated by native speakers of each language combination. In $\datasetname$, the covered language combinations include five combinations: Indonesian–English (ID-EN), Sundanese–Indonesian–English (SU-ID-EN), Javanese–Indonesian–English (JV-ID-EN), Hausa–English (HA-EN), and Algerian Arabic–Standard Arabic–French (AR-DZ-FR). This design ensures both linguistic fidelity and overall resource quality. Consequently, our benchmark provides a more reliable evaluation setting compared to datasets generated automatically by LLMs.

\subsection{Annotator Hiring}
We initiate the annotator recruitment process by first identifying the most prevalent language combinations exhibiting code-switching phenomena. For each language combination, we recruit a language champion---a native speaker responsible for managing data collection for that combination. The language champions then coordinate the recruitment of annotators to contribute to dataset construction.

\subsubsection{Recruitment Form}
Each language champion prepares a recruitment form. This form serves two primary purposes: (i) collecting demographic information from prospective annotators, and (ii) filtering respondents into the final pool of selected annotators. The form is structured into several sections, including an introduction, self-assessment, and language assessment. Further details regarding the recruitment form and collected annotator demographics are provided in Appendix~\ref{sec:appendix_recruitment_form} and ~\ref{sec:appendix_annotator_demographics}, respectively.

\subsubsection{Grouping}
For each selected annotator, we organize dialogue collection groups using the Discord\footnote{\url{https://discord.com/}.} platform. Given the limited number of recruited annotators (detailed guidelines are provided in Appendix~\ref{sec:appendix_annotator_guideline}), each individual participates in multiple groups. In total, we establish 100 dialogue collection groups for each language combination, comprising 50 groups with two speakers, 25 groups with three speakers, and 25 groups with four speakers. To support this process, we notify group members at three key points: five minutes before the session, at the start of the session, and at its conclusion. Annotators who are not directly involved in dialogue collection are subsequently assigned to annotate the downstream task datasets.

\subsection{Dataset Creation}

\subsubsection{Conversational Dialogue Collection}

We conduct a separate session for each assigned annotation group. In each session, participants engage in a 15-minute conversation on a pre-assigned topic. To capture natural code-switching behavior, participants are encouraged to mix languages within the designated language combinations whenever it feels natural to do so. Code-switching is allowed at multiple levels, including individual words, sentences, paragraphs, or even within a single utterance. To preserve anonymity, participants are asked not to refer to one another by name, whether in full, abbreviated, or nickname form. Instead, they are instructed to use Discord's mention feature whenever direct reference to another participant is necessary. This procedure ensures that all participant identities remain anonymized upon data export. Additionally, participants are allowed to use Discord's reply feature to respond directly to specific utterances, thereby maintaining a coherent dialogue structure.

\subsubsection{Question Answering}

Once the dialogues are collected, annotators write up to 10 question-answering items for each of the corresponding dialogue, these consisted of up to 5 answerable questions and 5 unanswerable questions. The answerable questions were designed to emphasize reasoning, such that answers could not be obtained directly from the dialogue. Instead, they required the use of external knowledge (primarily knowledge specific to particular cultural regions or discussed topics, where people who lived there or familiar with the topic would be familiar) and structured reasoning to reach a correct answer. The unanswerable questions, on the other hand, were constructed following five categories---Negation, Antonym, Entity--Swap, Mutual-Exclusion, and Impossible Condition---as defined in SQuAD 2.0 or SQuADRUn~\cite{rajpurkar-etal-2018-know}. All QA items were formulated as multiple-choice questions with five options, where option \textit{E} explicitly denoted ``\textit{No correct answer}''. Furthermore, each question was written in the designated first language (L1) of the corresponding language combination. For example, for the Javanese--Indonesian--English dataset, Indonesian was used as the language for the QA construction.

\subsubsection{Dialogue Summarization}

For each dialogue, we collect up to 3 distinct summaries, each from different annotators. Each summary is 3-5 sentences long and written in the designated L1 of the corresponding language combination. Annotators were provided with the initial topic of the dialogue and instructed to only summarize relevant information, excluding any off-topic segments. We encouraged annotators to follow the four dimensions of summarization quality defined in \cite{kryscinski-etal-2019-neural}: \textbf{Coherence}---The logical flow and collective quality of all sentences in the summary, \textbf{Fluency}---The grammatical quality of each individual sentence, \textbf{Relevance}---The inclusion of only important, on-topic information, and \textbf{Consistency}---The degree to which all facts in the summary are supported by the source document. Table~\ref{tab:iaa_bertscore_f1} presents the pairwise inter-annotator agreement scores for the written summaries, measured using BERTScore F1.

\subsubsection{Topic Classification}

Since the dialogue were collected based on a pre-defined initial topic, therefore naturally we can convert our dataset into a topic classification one as well. Specifically, we map the initial topics into classes as follow: \textbf{Science/Technology} (e.g., scientific breakthroughs, research, new technology, etc.), \textbf{Entertainment} (Sports, Music, Tourism, and any other form of entertainments), \textbf{Social/Culture} (e.g., languages, work culture, customs, traditions, etc.) \textbf{Education} (e.g., curriculum, schools, colleges, etc.), and \textbf{Daily Life} (e.g., personal experiences, love stories, daily habits). Table~\ref{tab:dialogue_topics_distribution} shows the distribution of the topics in the dataset.

\begin{table}[!t]
    \centering
    \setlength{\abovecaptionskip}{3pt}
    \resizebox{0.49\textwidth}{!}{
    \begin{tabular}{lccc}
    \toprule
        \textbf{Lang} & \textbf{Annotator 1--2} & \textbf{Annotator 1--3} & \textbf{Annotator 2--3} \\
    \midrule
        \texttt{ID--EN}    & 0.884 & 0.891 & 0.882 \\
        \texttt{JV--ID--EN} & 0.891 & 0.888 & 0.884 \\
        \texttt{SU--ID--EN} & 0.885 & 0.895 & 0.886 \\
        \texttt{ARQ--FRA}  & 0.858 & --    & --    \\
        \texttt{HA--EN}    & 0.883 & 0.887 & 0.880 \\
    \bottomrule
    \end{tabular}
    }
    \caption{Pairwise BERTScore F1 between annotator-written summaries as an inter-annotator agreement score across different language combinations (Lang).}  
    \label{tab:iaa_bertscore_f1}
\end{table}

\begin{table}[!t]
    \centering
    \setlength{\abovecaptionskip}{3pt}
    \resizebox{0.49\textwidth}{!}{
    \begin{tabular}{lcccccccc}
    \toprule
        \textbf{Lang} & \textbf{\#Dialogue} & \textbf{\#Sci/Tech} & \textbf{\#Ent} & \textbf{\#Soc/Cul} & \textbf{\#Edu} & \textbf{\#Daily} \\
    \midrule
        \texttt{ID--EN}    & 100 & 13 & 21 & 32 & 15 & 19 \\
        \texttt{JV--ID--EN} & 100 & 4 & 17 & 22 & 17 & 40 \\
        \texttt{SU--ID--EN} & 100 & 7 & 18 & 48 & 9 & 18 \\
        \texttt{HA--EN}    & 100 & 4 & 15 & 25 & 5 & 51 \\
        \texttt{AR--DZ--FR} & 100 & 14 & 21 & 21 & 24 & 20 \\
    \bottomrule
    \end{tabular}
    }
    \caption{Topic distribution of the different dialogues per language combination (Lang). We show the total number of dialogues (\#Dialogue) and per topic---science and technology (\#Sci/Tech), entertainment (\#Ent), society and culture (\#Soc/Cul), education (\#Edu), and daily life (\#Daily).}    
    \label{tab:dialogue_topics_distribution}
\end{table}

\subsection{Data Statistics}

The statistics of dialogues in $\datasetname$ are summarized in Table~\ref{tab:dialogue_data_size}. Our dataset comprises long, spontaneous conversations between 2 to 4 participants, providing a representative sample of authentic human interactions. As such, it is uniquely suited for benchmarking models on long-context, natural code-switching behavior. To quantify the linguistic complexity of these dialogues, we report two standard metrics: \textbf{Code-Mixing Index (CMI)} and \textbf{Switch Point Fraction (SPF)}.

\paragraph{Code-Mixing Index (CMI).} CMI \cite{gamback2016comparing} measures the intensity of code-switching by assessing the distribution of languages within an utterance. While the original formulation suggests treating Named Entities separately to avoid an artificial inflation of the index, robust NER tools are often unavailable for the under-resourced languages in our study. Consequently, following \citet{winata2019code}, we simplify the calculation by not distinguishing NEs.

\paragraph{Switch Point Fraction (SPF).} While CMI captures the volume of mixing, SPF \cite{pratapa2018language} focuses on the frequency of transitions between languages. It is calculated as the ratio of the number of language-switch points to the total number of possible switching positions.

\begin{figure}[!t]
  \centering
  \setlength{\abovecaptionskip}{3pt}
  \includegraphics[width=0.49\textwidth]{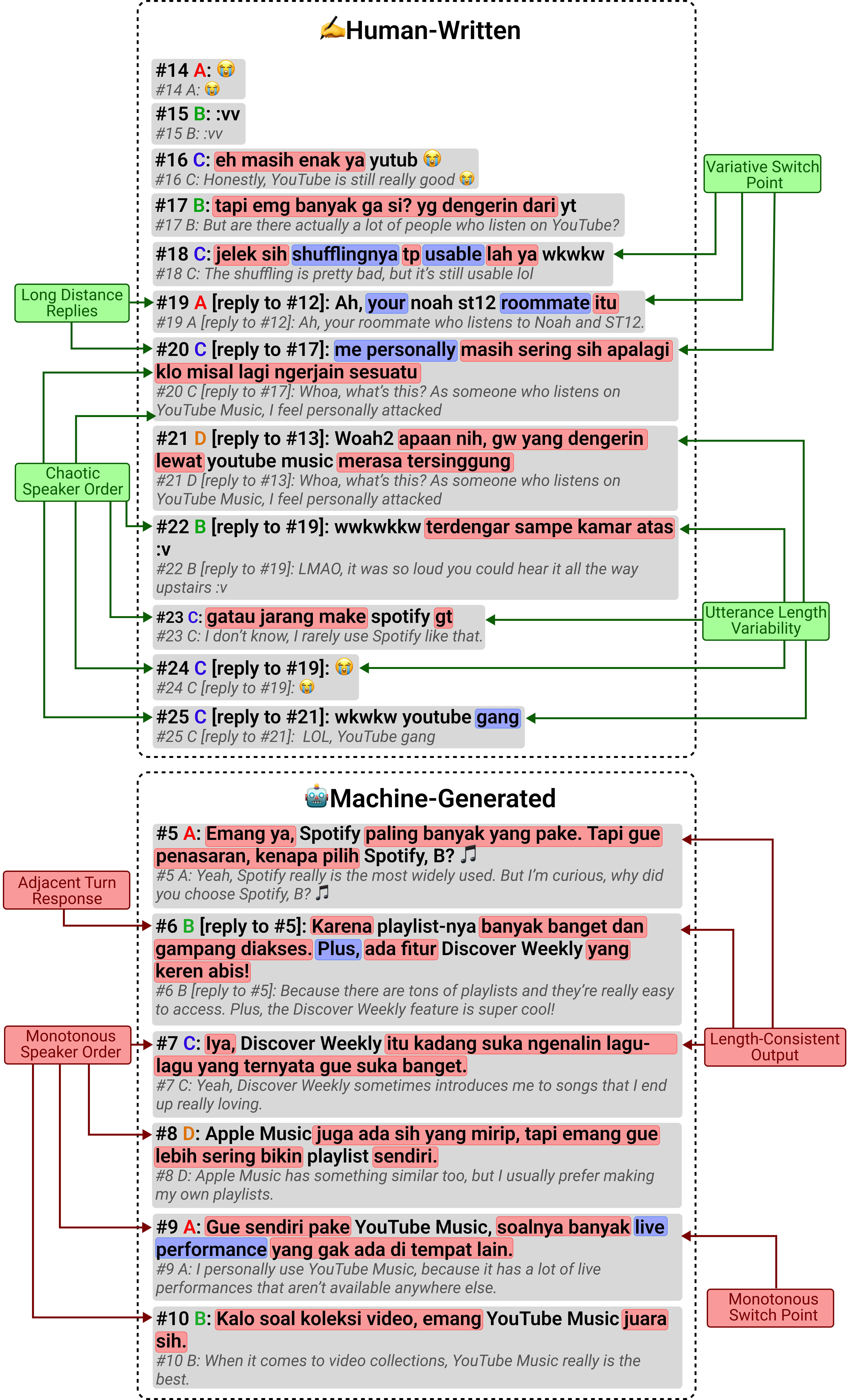}
    \caption{Comparison between human-written and machine-generated conversation texts. Text with a \colorIndicator{Black} color denotes the original conversation, while text with a \colorIndicator{Gray} color represents the English translation. Segments highlighted in \colorIndicator{Red} indicate Indonesian words, whereas segments highlighted in \colorIndicator{BlueThirty} indicate English words.}
  \label{fig:human_vs_machine}
\end{figure}

\begin{table*}[t]
    \centering
    \normalsize
    \small
    \setlength{\abovecaptionskip}{3pt}
    \resizebox{0.85\textwidth}{!}{
    \begin{tabular}{lcccccccc}
    \toprule
        \textbf{Lang.} & \textbf{\#Dialogue} & \textbf{Avg. Turn} & \textbf{Avg. Words} & \textbf{Avg. Tokens} & \textbf{Avg. CMI} & \textbf{Avg. SPF} \\
    \midrule
        \texttt{ID--EN}    & 100 & 81.93 & 448.96 & 782.15 & 0.472 & 0.300 \\
        \texttt{JV--ID--EN} & 100 & 83.07 & 410.14 & 754.04 & 0.467 & 0.318 \\
        \texttt{SU--ID--EN} & 100 & 59.88 & 494.03 & 957.20 & 0.757 & 0.467 \\
        \texttt{HA--EN}    & 100 & 69.14 & 421.32 & 649.84 & 0.352 & 0.190 \\
        \texttt{AR--DZ--FR} & 100 & 98.60 & 462.74 & 1,120.81 & 0.568 & 0.253 \\
    \bottomrule
    \end{tabular}
    }
    \caption{\textbf{Dialogue data statistics.} For each language combination (Lang.), we report the average number of turns (Avg. Turn), average number of words (Avg. Words), average number of tokens (Avg. Tokens), average Code-Mixing Index (Avg. CMI), and average Switch-Point Fraction (Avg SPF) per dialogue.}
    \label{tab:dialogue_data_size}
\end{table*}

\subsection{Natural Conversational Pattern}
Our dataset reflects the organic characteristics of human text-based communication, particularly in multi-party settings. We observe significant participation imbalance, where certain speakers dominate the conversation while others remain less active. Furthermore, human dialogue often exhibits multi-threaded structures; speakers frequently reply to refer and respond to messages from several turns prior. Utterance length is also highly variable, ranging from detailed explanations to brief one- or two-word reactions. Notably, human participants sometimes send multiple consecutive messages before a turn change occurs.

\begin{table}[!th]
    \small
    \centering
    \setlength{\abovecaptionskip}{3pt}
    \resizebox{0.49\textwidth}{!}{
    \begin{tabular}{@{}lcc@{}}
    \toprule
    \textbf{Metric} & \textbf{Human-written} & \textbf{Machine-generated} \\
    \midrule 
    \textbf{2 speakers (50 dialogue)} \\
    \midrule
     Avg length variance (tokens) & 66.540 & 35.140 \\
     Total replies & 658.2 & 2.6 \\
     Avg degree of reply distance & 2.729 & 0.048 \\
     Avg imbalance ratio of speaker turns & 1.366 & 1.019 \\
     Avg CMI & 0.525 & 0.541 \\
     Avg SPF & 0.306 & 0.313 \\
     Human preference & 2.727 & 2.178 \\
     \midrule 
    \textbf{3 speakers (25 dialogue)} \\
    \midrule
     Avg length variance (tokens) & 59.093 & 30.230 \\
     Total replies & 630.4 & 19.8 \\
     Avg degree of reply distance & 3.698 & 0.738 \\
     Avg imbalance ratio of speaker turns & 2.961 & 1.056 \\
     Avg CMI & 0.521 & 0.486 \\
     Avg SPF & 0.305 & 0.284 \\
     Human preference & 2.720 & 2.141 \\
     \midrule 
    \textbf{4 speakers (25 dialogue)} \\
    \midrule
     Avg length variance (tokens) & 81.195 & 29.191 \\
     Total replies & 869.6 & 45.0 \\
     Avg degree of reply distance & 4.159 & 1.085 \\
     Avg imbalance ratio of speaker turns & 3.256 & 1.118 \\
     Avg CMI & 0.520 & 0.483 \\
     Avg SPF & 0.306 & 0.284 \\
     Human preference & 2.676 & 2.078 \\
    \bottomrule
    \end{tabular}
    }
    \caption{Comparison of conversational statistics between human-written and machine-generated dialogues, averaged across all language combinations. Human preference scores are measured using a 1--3 Likert scale.}
\label{tab:natural_conversational_pattern}
\end{table}


In contrast, machine-generated dialogues produced by GPT-4o based on the same topics appear significantly more monotonous. These conversations tend to follow a rigid turn-taking structure with a uniform distribution among speakers and consistent utterance lengths. Synthetic dialogues rarely feature consecutive messages from a single speaker, and replies are almost always linear and immediate. A comparison of these patterns is illustrated in Figure \ref{fig:human_vs_machine}. To quantify these observations, we provide statistical analysis in Table \ref{tab:natural_conversational_pattern}, which shows that across most languages, human annotators consistently rate organic conversations as more natural than their machine-generated counterparts. We use Free-marginal multirater kappa~\cite{randolph2010kappa} to calculate the inter-annotator agreement, obtaining a pooled score of 0.172. This slight agreement reflects not only the subjective nature of evaluating naturalness, but also the high variance across distinct language combinations. For example, in 2-speaker settings, the human-over-machine preference gap is large for AR-DZ-FR (2.94 vs 1.21) and SU-ID-EN (2.84 vs 2.12), yet remarkably small for ID-EN (2.72 vs 2.69) and HA-EN (2.48 vs 2.47). A pooled $\kappa$ inevitably understates agreement when averaging over populations this heterogeneous. Detailed statistics for each language combination are provided in Appendix~\ref{sec:quantitative-results-human-vs-machine}.

\begin{table*}[!th]
\centering
\setlength{\abovecaptionskip}{3pt}
\resizebox{\textwidth}{!}{
\begin{tabular}{@{}lccccccccccccccc@{}}
\toprule
& \multicolumn{5}{c}{\textsc{Question Answering (Accuracy \%)}} & \multicolumn{5}{c}{\textsc{Dialogue Summarization (BERTScore-F)}} & \multicolumn{5}{c}{\textsc{Topic Classification (Accuracy \%)}}\\
\cmidrule(lr){2-6} \cmidrule(lr){7-11} \cmidrule(lr){12-16}
\textbf{Model} & 
    \texttt{\textbf{ID--EN}} & 
    \texttt{\textbf{JV--ID--EN}} & 
    \texttt{\textbf{SU--ID--EN}} & 
    \texttt{\textbf{HA--EN}} & 
    \texttt{\textbf{AR--DZ--FR}} & 
    \texttt{\textbf{ID--EN}} & 
    \texttt{\textbf{JV--ID--EN}} & 
    \texttt{\textbf{SU--ID--EN}} & 
    \texttt{\textbf{HA--EN}} & 
    \texttt{\textbf{AR--DZ--FR}} &  
    \texttt{\textbf{ID--EN}} & 
    \texttt{\textbf{JV--ID--EN}} & 
    \texttt{\textbf{SU--ID--EN}} & 
    \texttt{\textbf{HA--EN}} & 
    \texttt{\textbf{AR--DZ--FR}} \\
\midrule
\textbf{Global} \\
\midrule
Qwen2.5-3B-Instruct & 5.86 & 18.98 & 76.16 & 43.93 & 22.22 & 0.894 & 0.893 & 0.901 & 0.777 & 0.784 & 52.53 & 38.38 & 52.53 & 30.30 & 46.46 \\
Qwen2.5-7B-Instruct & 4.64 & 18.98 & 88.68 & 43.93 & 38.38 & 0.886 & 0.885 & 0.895 & 0.822 & 0.812 & 54.55 & 60.61 & 47.47 & 60.61 & 56.57 \\
Qwen3-4B & 8.88 & 30.90 & 90.70 & 49.49 & 39.39 & 0.887 & 0.888 & 0.900 & 0.748 & 0.802 & 52.53 & 46.46 & 44.44 & 44.44 & 49.49 \\
Qwen3-8B & 5.65 & 31.71 & 89.89 & 41.41 & 50.00 & 0.879 & 0.873 & 0.892 & 0.754 & 0.774 & 54.55 & 56.57 & 47.47 & 49.49 & 55.56 \\
Aya23-8B & 8.28 & 23.03 & 80.00 & 28.78 & 14.14 & 0.896 & 0.894 & 0.902 & 0.803 & 0.786 & 57.58 & 47.47 & 49.49 & 57.58 & 46.46 \\
Gemma2-9B-Instruct & 9.09 & 39.19 & 91.51 & 81.81 & 49.49 & 0.903 & 0.898 & 0.908 & 0.879 & 0.812 & 60.61 & 42.42 & 50.51 & 56.57 & 39.39 \\
Gemma3-4B-Instruct & 23.43 & 33.93 & 90.50 & 62.62 & 38.38 & 0.872 & 0.857 & 0.875 & 0.827 & 0.764 & 41.41 & 42.42 & 44.44 & 57.58 & 34.34 \\
\midrule
\textbf{Regional} \\
\midrule
Sailor2-8B & 27.67 & 40.40 & 86.46 & - & - & 0.736 & 0.660 & 0.588 & - & - & 45.45 & 49.49 & 31.31 & - & - \\
Sahabat-AI-Gemma & 13.73 & 43.23 & 93.53 & - & - & 0.899 & 0.895 & 0.907 & - & - & 62.63 & 53.54 & 44.44 & - & - \\
SILMA-9B-Instruct & - & - & - & - & 0.0 & - & - & - & - & 0.733 & - & - & - & - & 34.34 \\
ALLAM-7B-Instruct & - & - & - & - & 32.32 & - & - & - & - & 0.359 & - & - & - & - & 34.34 \\
\bottomrule
\end{tabular}
}
\caption{\textbf{Experimental Results on $\datasetname$.} This ablation is conducted under a zero-shot prompting setting, with the reasoning (thinking) mode disabled for models that support it. For the Question Answering task, we report results on the answerable subset.}
\label{tab:ablation_experiment_result}
\vspace{-1em}
\end{table*}

\section{Experimental Setup}

We evaluate a diverse set of models across the three tasks. From the perspective of language coverage, these models can be categorized into English-centric models, multilingual models, and region-specific models tailored to particular language combinations. We also compare base models with reasoning-oriented and instruction-tuned variants.

Regarding linguistic coverage, our study leverages models such as \textsc{Sailor2}~\cite{dou2025sailor2}, \textsc{Aya23}~\cite{aryabumi2024aya}, \textsc{Sahabat AI},\footnote{\url{https://huggingface.co/Sahabat-AI}.} \textsc{Qwen2.5}~\cite{qwen2025qwen25technicalreport}, \textsc{Gemma2}~\cite{team2024gemma}, \textsc{Gemma3}~\cite{team2025gemma}, \textsc{Allam}~\cite{bari2025allam}, and \textsc{Silma}~\cite{silma-9b-2024}. Furthermore, to assess reasoning capabilities, we include \textsc{Qwen3}~\cite{yang2025qwen3} models with 4B and 8B parameters. Detailed hyperparameter configurations for each model are provided in Appendix~\ref{tab:appendix_hyperparameter_settings}. We do single-run for our experiment due to the cost. Moreover, our novelty is in the dataset and we do not claim definite ranking on the models, rather we show general performance across diverse set of models and setting in our dataset.

\subsection{Reasoning Behavior}

Some of our models, namely the Qwen3 series, have reasoning capabilities. Therefore, we investigate whether thinking trace is beneficial in our dataset. For models without built-in thinking behavior, we can elicit reasoning by explicitly asking the model to generate their reasoning trace first, akin to prior work such as chain-of-thought~\cite{wei2022chain}. 

\subsection{Task Setup}

Across all downstream tasks, we adopt a set of shared experimental prompt configurations, which include the construction of example dialogues, the number of shots, and the specification of output formats. These configurations are further adapted to the requirements of each individual task. The prompts are detailed in Appendix~\ref{sec:appendix_prompt_example}.

\paragraph{Number of Shots.}

For QA, we experiment with 0 and 1 shot prompting. For Dialogue Summarization, we use 0, 1, and 3-shot settings of summary examples without a dialogue example. The in-context examples for each task are drawn from dialogues that are not part of the evaluation set.



\subsection{Evaluation Metrics}

All models are assessed with task-specific evaluation metrics. For \textsc{Question Answering} and \textsc{Topic Classification} tasks, we use accuracy as the evaluation metric. For \textsc{Dialogue Summarization} task, we use ROUGE \cite{lin-2004-rouge}, METEOR \cite{banerjee-lavie-2005-meteor}, CHRF++ \cite{popovic-2015-chrf}, and BERTScore \cite{zhang2020bertscoreevaluatingtextgeneration}, as implemented in the Hugging Face \texttt{evaluate} library. These metrics were chosen because they have a high correlation with the four dimensions of summarization quality, according to \cite{fabbri-etal-2021-summeval}. 
In the main paper we report BERTScore-F, but other metrics are shown in Appendix~\ref{sec:quantitative-results-summary}.

\section{Results and Analysis}


The performance metrics for our evaluated models across all target languages are summarized in Table~\ref{tab:ablation_experiment_result} (Full results are detailed in Appendix~\ref{sec:quantitative_result}). A key takeaway from these results is that the majority of models exhibit poor performance, which suggests that the proposed benchmark represents a significant challenges for current systems. Notably, we observe a a performance gap between general-purpose multilingual models and regionally-designed ones. The latter generally achieve better results, highlighting the clear benefits of utilizing specialized models that are tailored to specific linguistic and cultural characteristics.

While there is notable variance in performance across different language groups, it is difficult to conclude that any specific language is inherently more difficult than others. This is primarily due to the non-parallel nature of our dataset; the difficulty remains anchored to the specific conversation content of each language (see Appendix~\ref{sec:appendix_qualitative_analysis} and Appendix ~\ref{sec:qualitative_summarization}).

\begin{figure*}[!t]
  \centering
  \setlength{\abovecaptionskip}{3pt}
  \includegraphics[width=0.9\linewidth]{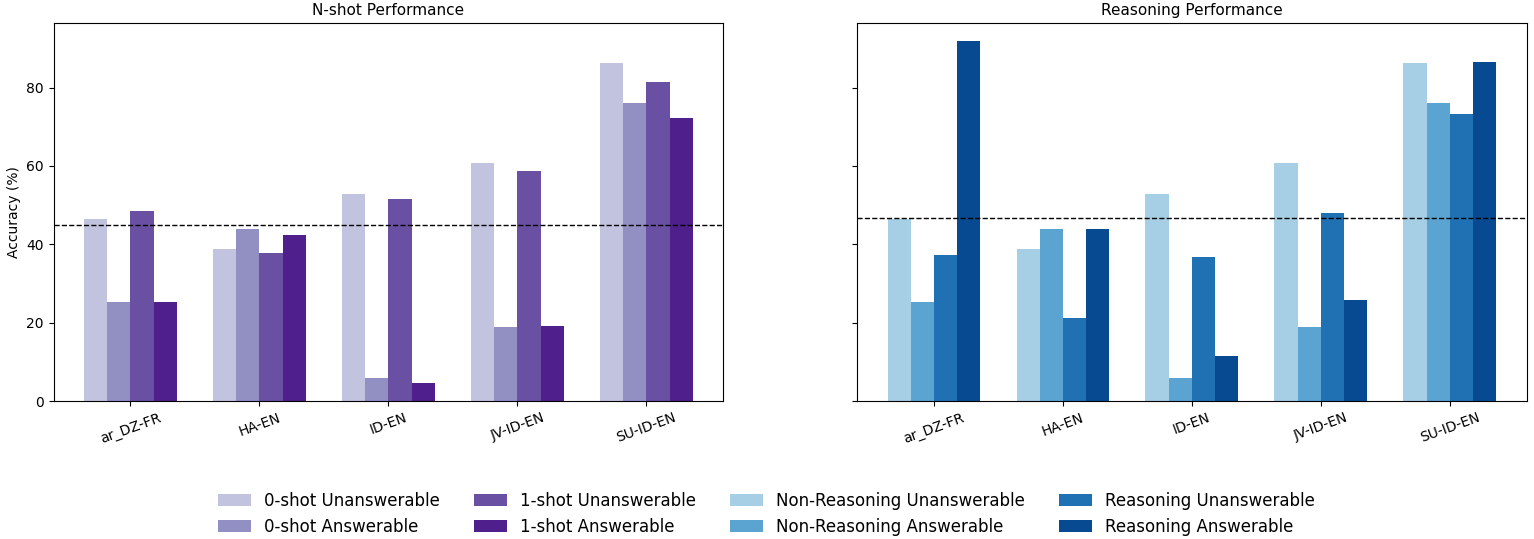}
    \caption{Comparison of average model performance (Acc.\ \%) across languages for Answerable vs.\ Unanswerable cases, from the perspectives of N-shot prompting (Left) and reasoning performance (Right).}
  \label{fig:answerable_vs_unanswerable}
\end{figure*}

\subsection{Effect of Reasoning}

\begin{figure}[!t]
  \centering
  \setlength{\abovecaptionskip}{3pt}
    \includegraphics[width=\linewidth]{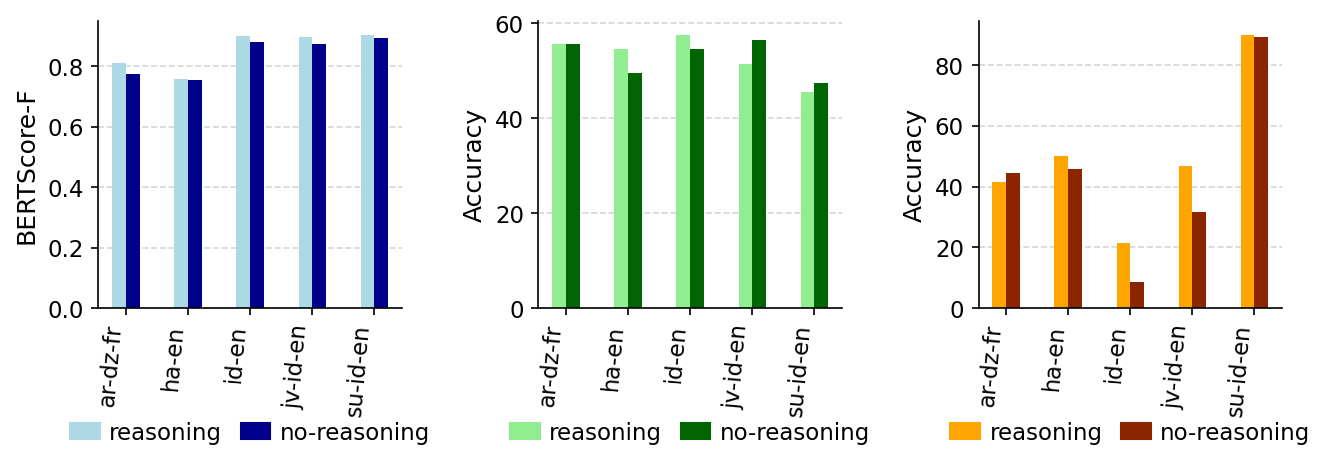}
    \caption{Reasoning vs No-Reasoning performance in Dialogue Summarization (blue), Topic Classification (green), and QA (red).}
    \label{fig:placeholder}
  \label{fig:reasoning-vs-noreasoning}
\end{figure}

We investigated the impact of explicit reasoning on model performance by utilizing the native ``thinking traces'' available in recent models like Qwen3, and by using explicit prompting to generate reasoning steps for other models. As shown in Figure~\ref{fig:reasoning-vs-noreasoning}, we observe a consistent improvement mostly in performance when reasoning is enabled. This indicates that the tasks within our dataset benefit from the additional computation and internal verification afforded by reasoning traces.

\subsection{Effect of Few-Shot Learning}

\begin{figure}[!th]
  \centering
  \setlength{\abovecaptionskip}{3pt}
  \includegraphics[width=\linewidth]{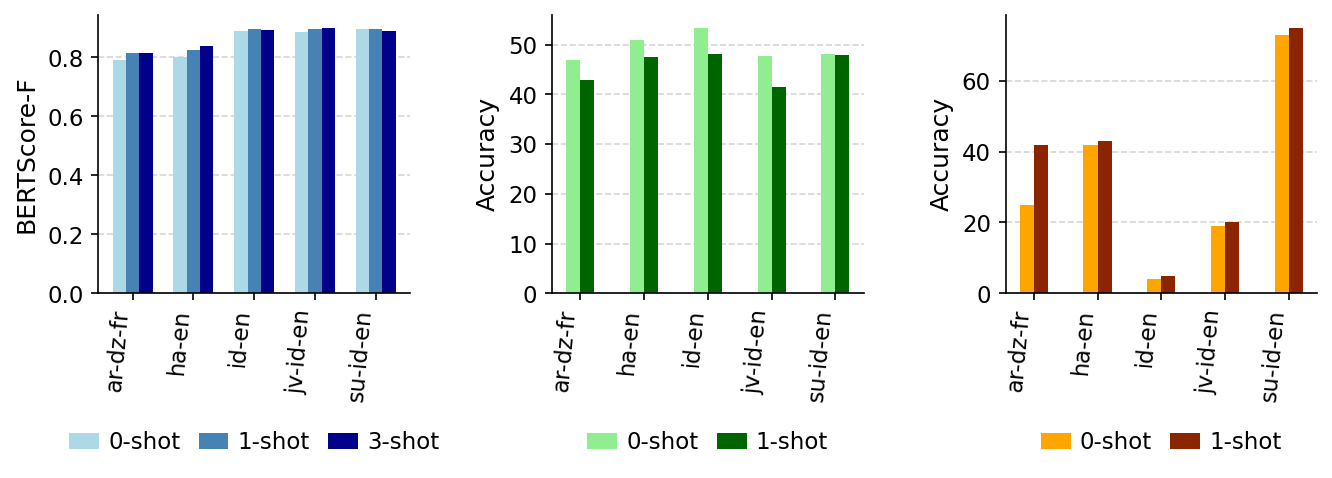}
    \caption{Zero-shot vs Few-shot performance in Dialogue Summarization (blue), Topic Classification (green), and QA (red).}
  \label{fig:effecy_of_few_shot}
\end{figure}

In comparison, the inclusion of few-shot examples does not appear to be a significant factor in improving model performance for QA and Topic Classification, but improves slightly in Dialogue Summarization. As shown in Figure~\ref{fig:effecy_of_few_shot}, providing a small number of in-context examples generally does not yield consistent gains across the evaluated models.

\subsection{Answerable vs. Unanswerable QA}

Based on the results in Figure~\ref{fig:answerable_vs_unanswerable}, we observe that answerable questions---specifically those designed to require explicit reasoning---exhibit consistent performance improvements when the model leverages reasoning mechanisms, whether implicitly or explicitly. In contrast, for the unanswerable subset, performance gains are observed only when explicit reasoning traces are enabled in the Qwen3 model, whereas implicit reasoning does not yield improvements and, in some cases, even leads to performance decline. This suggests that unanswerable cases place stricter demands on structured reasoning and error detection. Furthermore, a detailed analysis of shot settings indicates that few-shot prompting does not provide stable or uniform benefits across models. The effectiveness of few-shot examples appears to be model-dependent, with some models benefiting marginally while others show no improvement or even slight regressions.

\section{Conclusion}

In this work, we introduced $\datasetname$, a multi-party code-switching benchmark capturing the authentic complexity of human multilingual discourse across five language combinations. By utilizing human-authored conversations with 2 to 4 participants, we provide a dataset containing structural nuances, such as multi-threaded dynamics and varied speaker dominance, often missing in synthetic corpora. Our analysis confirms that these dialogues are more natural and structurally diverse than machine-generated alternatives, particularly in message length and reply distances. We build an evaluation pipeline that covers three downstream tasks based on the conversation: QA, Summarization, and Topic Classification to address gaps in prior benchmarks and reflect the requirements of the modern LLM era. Experimental results show that state-of-the-art models still struggle with the intricacies of natural, multi-party code-switching. This performance gap highlights the importance of \datasetname~in identifying current NLP limitations and serves as a foundation for developing more robust, inclusive systems for the world's multilingual majority.

\section*{Limitations}

\paragraph{Languages Coverage.} This work covers 7 complementary (under-studied) languages across 5 language combinations, spanning 3 geographic regions. While not exhaustive, this scope provides a scalable foundation for building natural code-switching datasets such as $\datasetname$, which can be readily extended to additional languages and regions. After completing the experiments, we extended the dataset to include the Hindi–English language pair. The results reported in this paper are based on the earlier version of the dataset and therefore do not cover this newly added pair. Due to practical constraints, the data cannot cover all regions and topics or reflect all cultural perspectives.

\paragraph{Models.} We evaluate a diverse set of global and regional LLMs that vary in linguistic coverage, model size, and reasoning capability, including both open-source and proprietary systems. As the LLM landscape continues to evolve, expanding this evaluation to a broader range of models represents a natural and promising direction for future research.

\paragraph{Non-Parallel Dataset.} Our dataset encompasses a wide range of code-switching language combinations, each annotated by native speakers of the corresponding language varieties under a unified annotation guideline. Consequently, the datasets and model performance measurements for each language setting are inherently independent rather than parallel, and therefore cannot be directly compared across languages. Nevertheless, this characteristic constitutes a key strength of our benchmark, as it preserves the naturalness and authenticity of the code-switching phenomena represented in the data.

\paragraph{Task Design.} In the Answerable Question subset of the QA task, the questions are intentionally designed such that they can only be solved through reasoning augmented with external knowledge, particularly local or cultural knowledge specific to a given region. As a result, several language combinations exhibit substantial performance degradation, in some cases falling below random-chance performance (<50\%). However, this property further demonstrates the challenging nature of our benchmark while simultaneously encouraging the development of fairer LLMs with more equitable representations of localized and culturally grounded knowledge across regions.

\paragraph{Evaluation Metric.} Finally, due to the lack of reliable tools for computing advanced CMI metrics in under-studied languages, we adopt a relaxed CMI formulation that preserves meaningful comparative insights. We expect that future advances in multilingual NLP resources will enable more fine-grained code-mixing analyses for these languages.

\section*{Ethical Considerations}
The annotators involved in this study were compensated above the local minimum wage. They received detailed instructions, and any demographic information collected was obtained with their informed consent (see Appendix~\ref{sec:appendix_recruitment_form}). Annotators were explicitly instructed not to use offensive language, and we conducted verification to the best of our ability. Nonetheless, some instances may have been missed, and we are committed to updating the benchmark if such cases are reported. While we aimed to construct realistic datasets, we do not claim that our benchmark captures all code-switching variations across the five language combinations. We also acknowledge the use of an AI assistant (e.g. ChatGPT, Gemini, Github Copilot) to support code implementation and to polish the writing of this manuscript.





\bibliography{custom}

\clearpage
\appendix

\section{Recruitment Form}
\label{sec:appendix_recruitment_form}

The recruitment form is organized into three main sections, as described below.

\begin{figure}[h]
  \centering
  \includegraphics[width=0.49\textwidth]{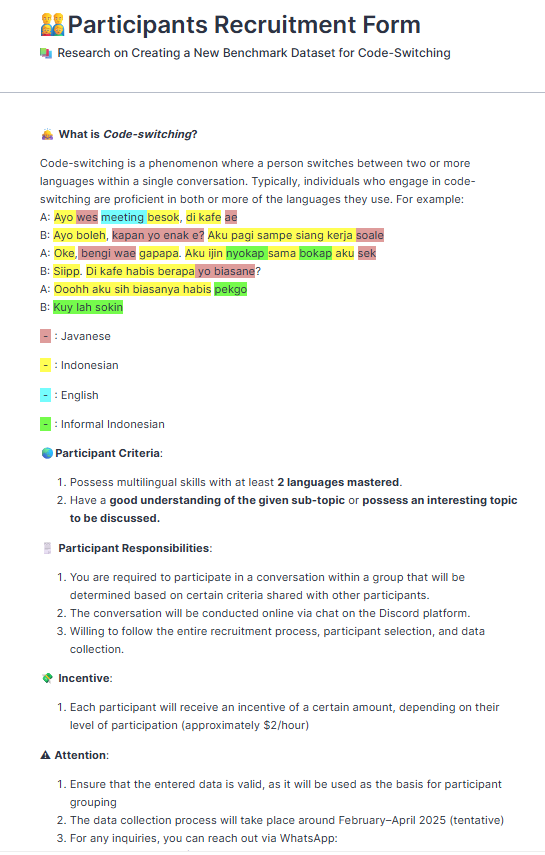}
  \caption{Introduction Section of the Recruitment Form.}
  \label{fig:recruitmentForm_introduction}
\end{figure}

\paragraph{Introduction.}
This section provides an overview and general background on code-switching, along with the eligibility criteria required for respondents to participate in the crowdsourcing process. The introduction section of the recruitment form is shown in Figure~\ref{fig:recruitmentForm_introduction}. This also shows what kind of data that will be collected from the participant in the process.

\begin{figure}[h]
  \centering
  \includegraphics[width=0.49\textwidth]{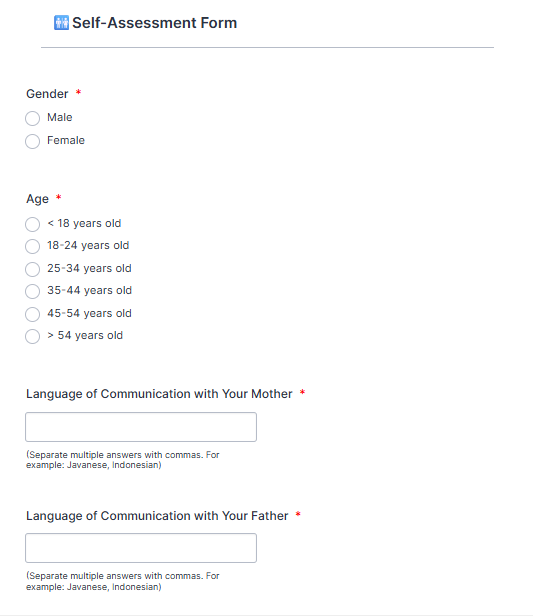}
  \caption{Self-Assessment Section (Part 1) of the Recruitment Form.}
  \label{fig:recruitmentForm_selfAssessment1}
\end{figure}

\begin{figure}[h]
  \centering
  \includegraphics[width=0.49\textwidth]{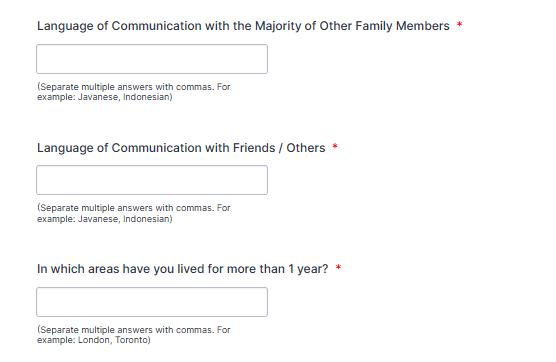}
  \caption{Self-Assessment Section (Part 2) of the Recruitment Form.}
  \label{fig:recruitmentForm_selfAssessment2}
\end{figure}

\begin{figure}[h]
  \centering
  \includegraphics[width=0.49\textwidth]{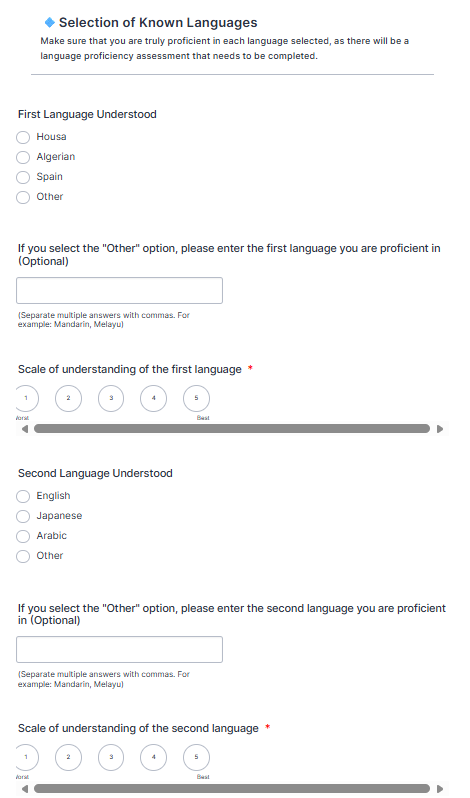}
  \caption{Self-Assessment Section (Part 3) of the Recruitment Form.}
  \label{fig:recruitmentForm_selfAssessment3}
\end{figure}

\paragraph{Self-Assessment.}
Respondents are asked to provide personal demographic information, as well as details about the languages they commonly use when communicating within their close social circles. The self-assessment section of the recruitment form is illustrated in Figures~\ref{fig:recruitmentForm_selfAssessment1}, Figure~\ref{fig:recruitmentForm_selfAssessment2}, and Figure~\ref{fig:recruitmentForm_selfAssessment3}.

\begin{figure}[h]
  \centering
  \includegraphics[width=0.49\textwidth]{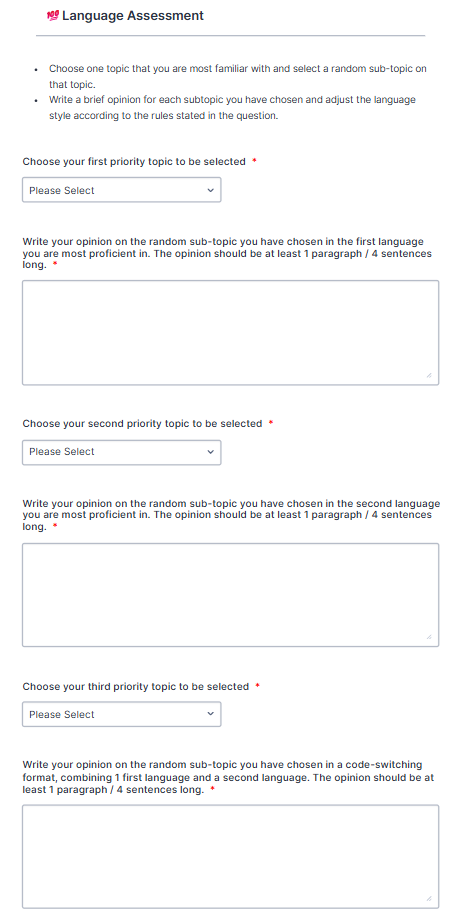}
  \caption{Language-Assessment Section of the Recruitment Form.}
  \label{fig:recruitmentForm_languageAssessment}
\end{figure}

\paragraph{Language Assessment.}
The recruitment form also includes a language assessment component, in which respondents are instructed to compose a short paragraph on a given topic using a specified combination of languages. This task is designed to simulate respondents’ natural code-switching behavior. The language assessment section of the recruitment form is shown in Figure~\ref{fig:recruitmentForm_languageAssessment}.

\section{Annotator Demographics}
\label{sec:appendix_annotator_demographics}

Table~\ref{tab:annotator_demographics_table} presents the demographic data of the annotators involved in the dataset creation.

\begin{table}[!th]
    \centering
    \resizebox{0.47\textwidth}{!}{
    \begin{tabular}{@{}lcccccccc@{}}
    \toprule
        & & \multicolumn{2}{c}{\textsc{Gender}} & \multicolumn{5}{c}{\textsc{Age (Years)}} \\
        \cmidrule(lr){3-4} \cmidrule(lr){5-9}
        \textbf{Lang} & \textbf{\#Annotators} & \textbf{Male} & \textbf{Female} & \textbf{18-24} & \textbf{25-34} & \textbf{35-44} & \textbf{45-54} & \textbf{55+} \\
    \midrule
        \texttt{ID--EN}    & 11 & 11 & 0 & 11 & 0 & 0 & 0 & 0 \\
        \texttt{JV--ID--EN} & 30 & 10 & 20 & 26 & 4 & 0 & 0 & 0 \\
        \texttt{SU--ID--EN} & 13 & 6 & 7 & 8 & 5 & 0 & 0 & 0 \\
        \texttt{HA--EN}    & 11 & 6 & 5 & 4 & 1 & 4 & 2 & 0 \\
        \texttt{AR--DZ--FR} & 11 & 7 & 4 & 1 & 9 & 0 & 0 & 1 \\
    \bottomrule
    \end{tabular}
    }
    \caption{Annotator Demographics Data}    
    \label{tab:annotator_demographics_table}
\end{table}

\section{Annotator Guideline}
\label{sec:appendix_annotator_guideline}

For each annotator-related task, a set of guidelines is provided to ensure that dataset construction is carried out consistently across languages.

\begin{figure}[h]
  \centering
  \includegraphics[width=0.49\textwidth]{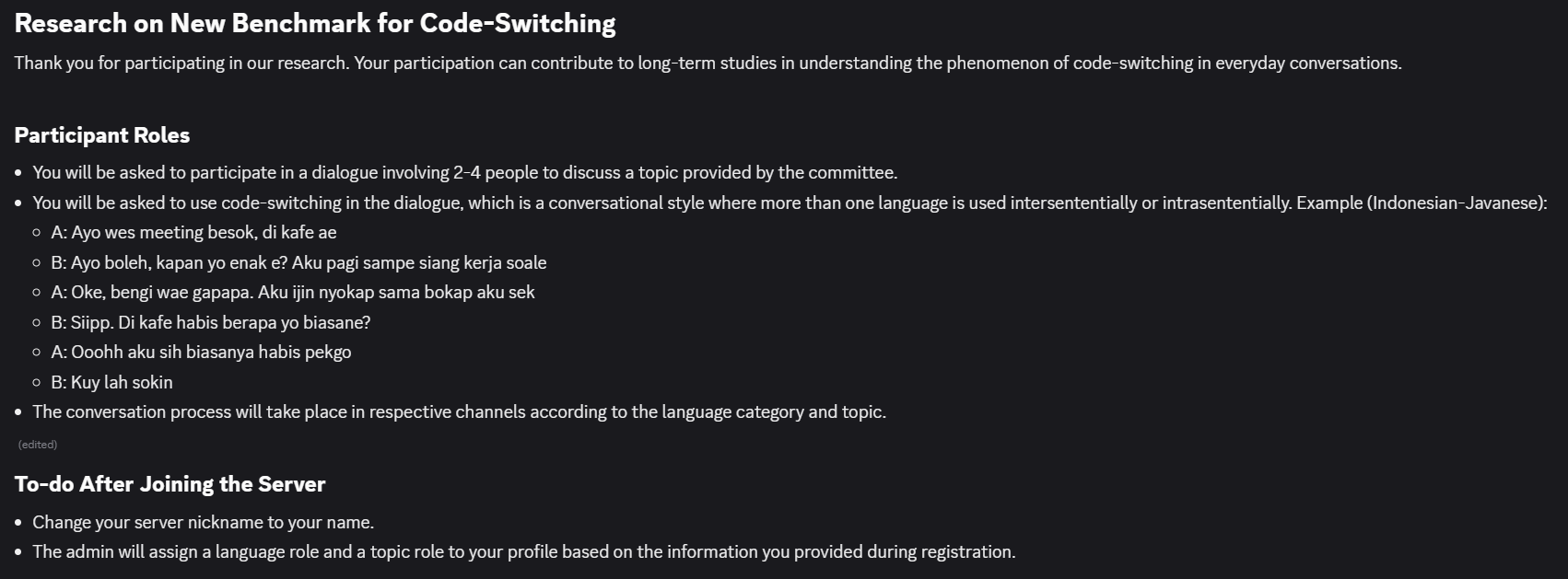}
  \caption{Dialog Construction Annotator Guideline (Part 1).}
  \label{fig:guideline_dialog_1}
\end{figure}

\begin{figure}[h]
  \centering
  \includegraphics[width=0.49\textwidth]{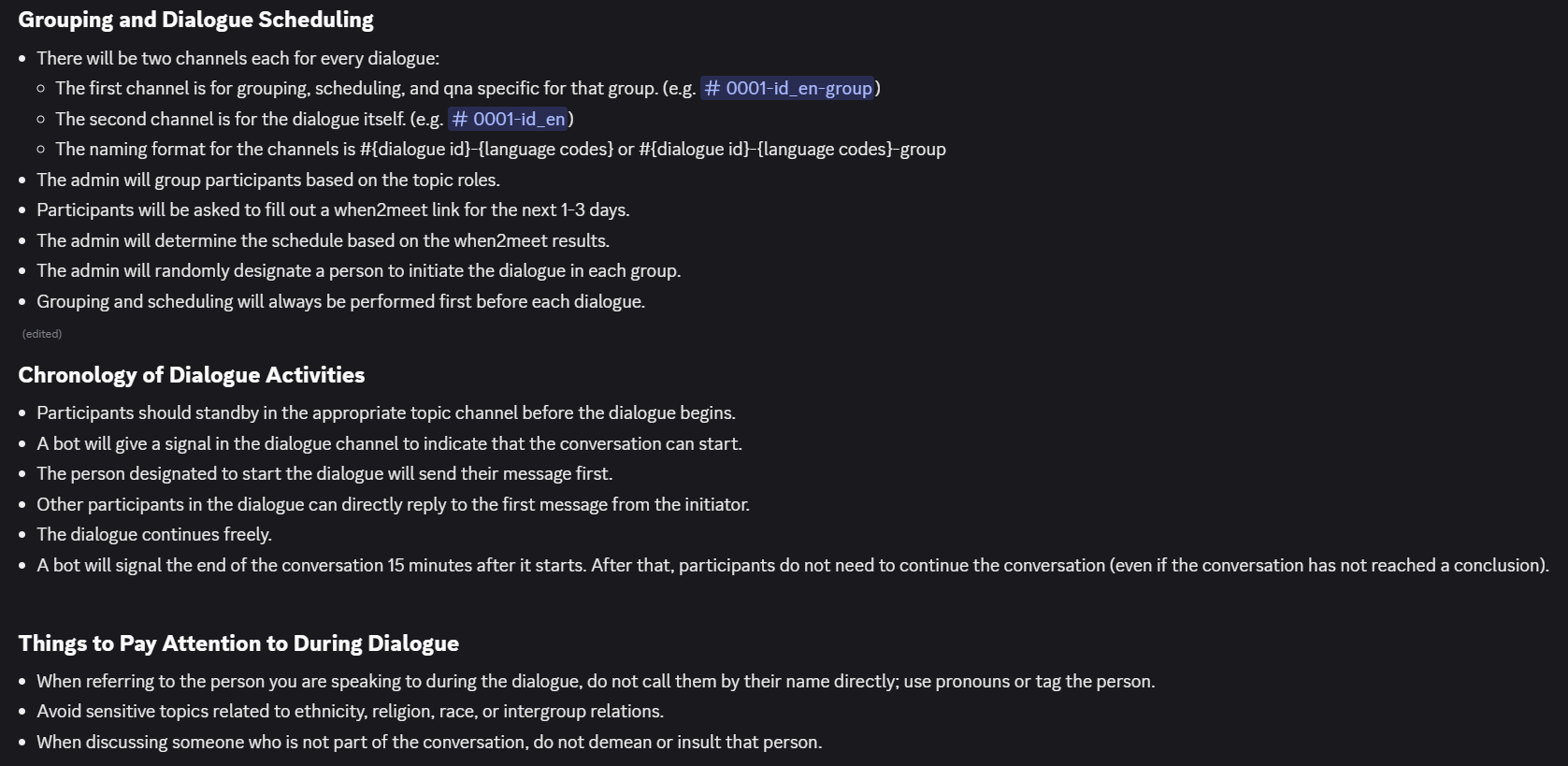}
  \caption{Dialog Construction Annotator Guideline (Part 2).}
  \label{fig:guideline_dialog_2}
\end{figure}

\paragraph{Dialogue Construction.}
Guidelines for dialogue construction are provided through a dedicated channel on the Discord platform, which is used for collecting the dialogue data. These guidelines describe the participants’ tasks, explain the grouping process, and specify the rules that must be followed during dialogue creation. The dialogue construction guidelines are shown in Figures~\ref{fig:guideline_dialog_1} and~\ref{fig:guideline_dialog_2}.

\begin{figure}[h]
  \centering
  \includegraphics[width=0.49\textwidth]{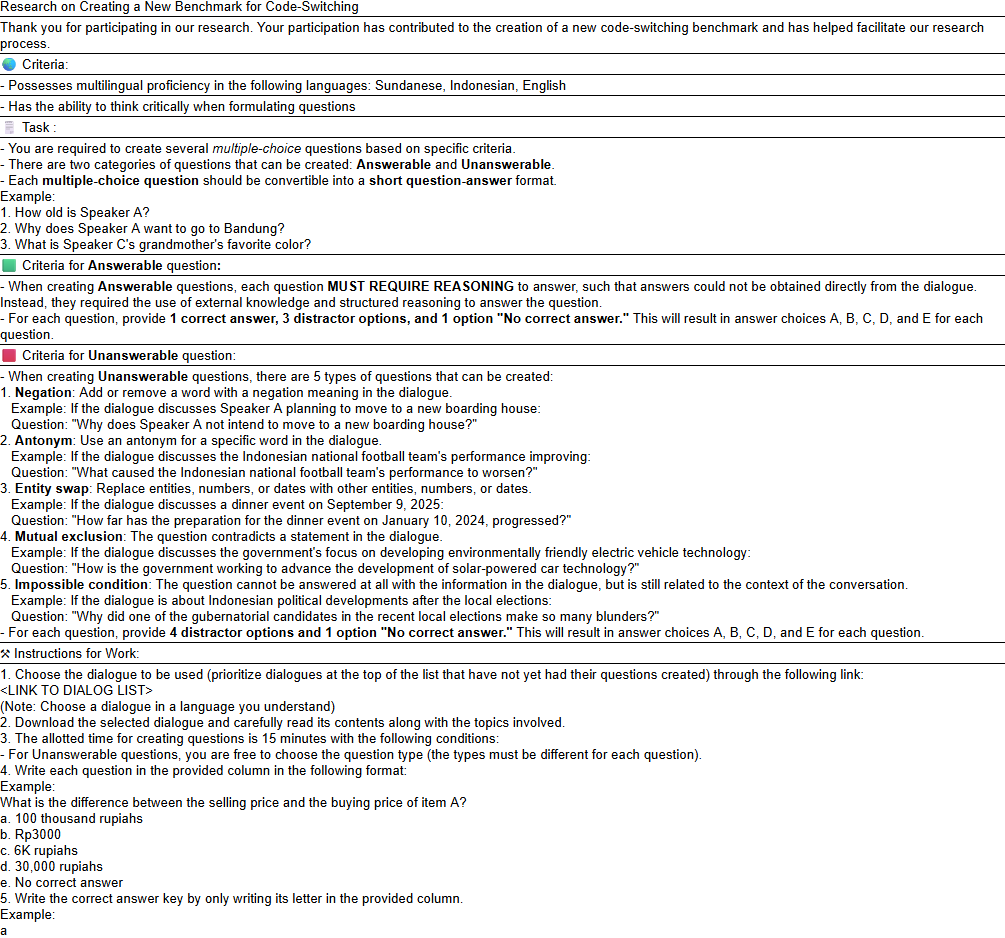}
  \caption{Question Answering Annotator Guideline.}
  \label{fig:guideline_qa}
\end{figure}

\paragraph{Question Answering.}
Guidelines for the Question Answering task are presented on the landing page of the annotation platform. These guidelines outline the annotators’ responsibilities, the types of questions that must be created, and examples of appropriate questions. The Question Answering guidelines are shown in Figure~\ref{fig:guideline_qa}.

\begin{figure}[h]
  \centering
  \includegraphics[width=0.49\textwidth]{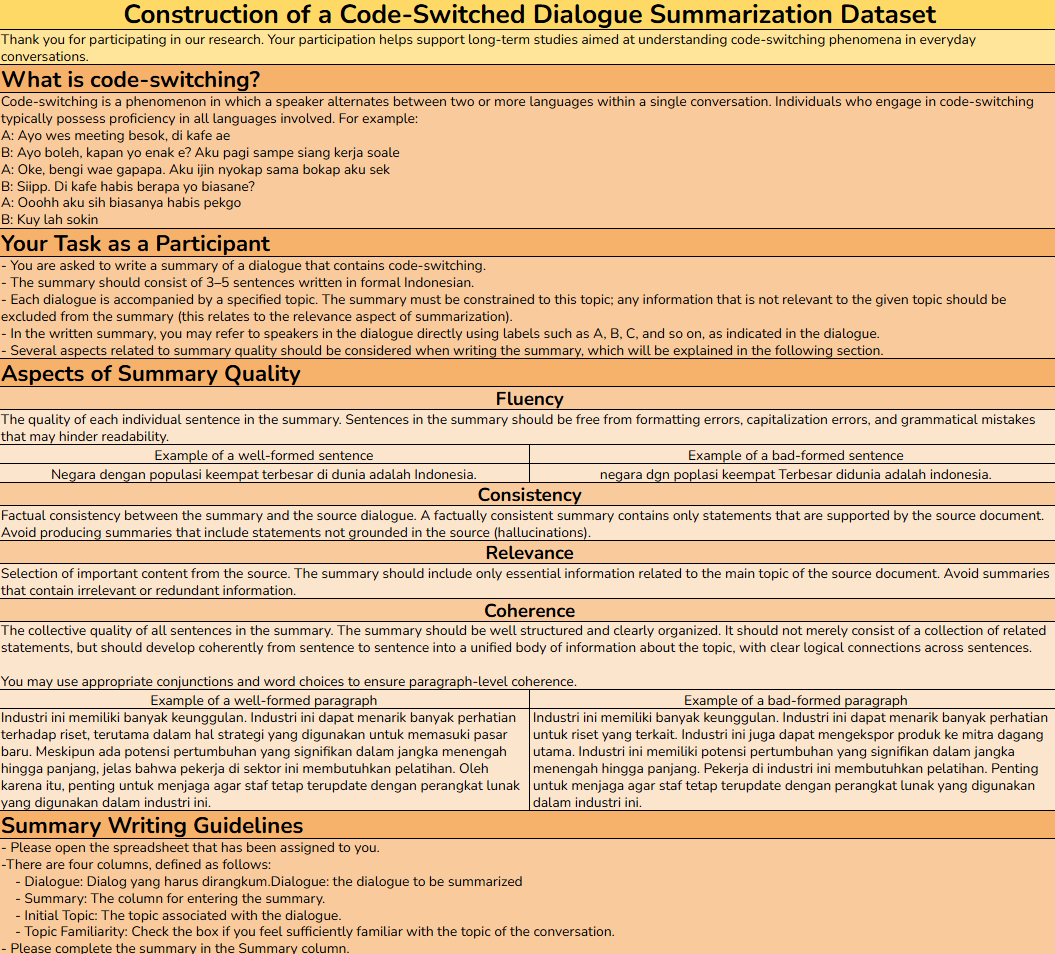}
  \caption{Dialogue Summarization Annotator Guideline.}
  \label{fig:guideline_summary}
\end{figure}

\paragraph{Dialogue Summarization.}
Guidelines for Dialogue Summarization are also provided on the landing page of the annotation platform. They describe the annotators’ tasks, as well as general criteria for producing high-quality summaries, accompanied by illustrative examples. The guidelines for this task are shown in Figure~\ref{fig:guideline_summary}.

\begin{figure}[h]
  \centering
  \includegraphics[width=0.49\textwidth]{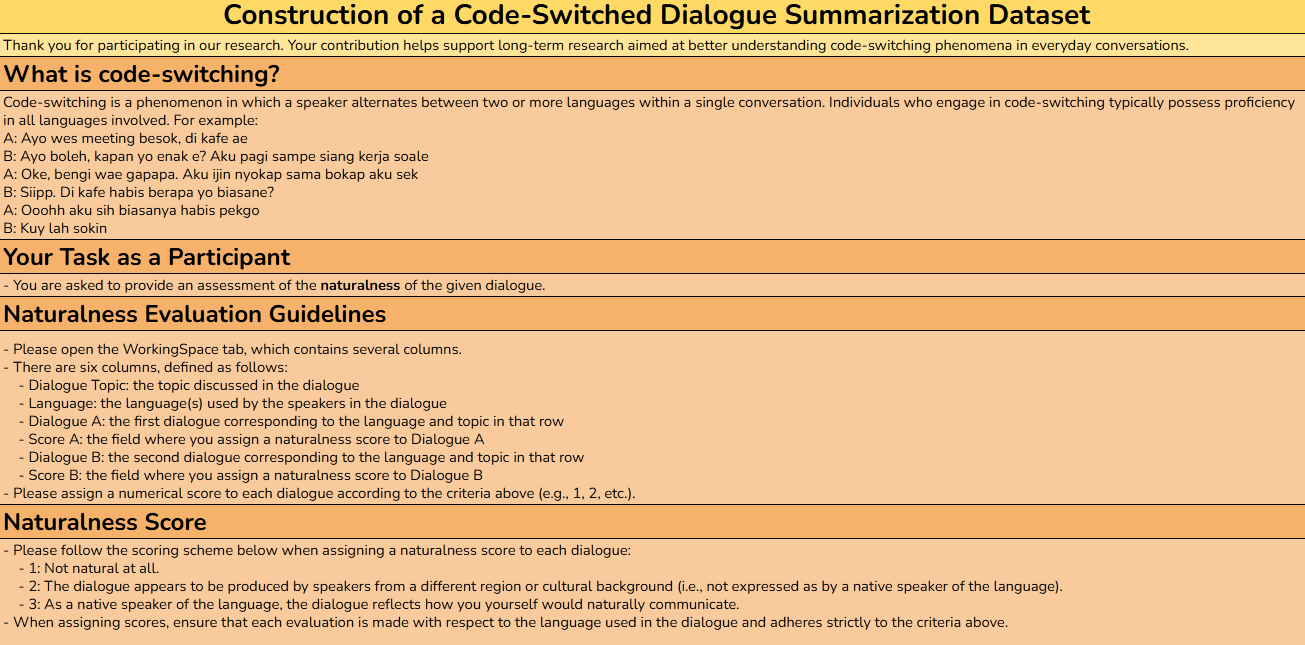}
  \caption{Human-written vs Machine-generated Naturalness Annotator Guideline.}
  \label{fig:guideline_naturalness}
\end{figure}

\paragraph{Naturalness.}
The Naturalness guidelines are presented in the same manner as those for Question Answering and Dialogue Summarization. This section specifies the annotators’ tasks and details the scoring criteria, which are based on prior work by \citet{yong-etal-2023-prompting}. The Naturalness guidelines are shown in Figure~\ref{fig:guideline_naturalness}.

\section{Prompt Example}
\label{sec:appendix_prompt_example}

The prompts used in our experiments consist of four components, as described below.

\paragraph{Question Answering.}
Figure~\ref{fig:prompt-question-answering} illustrates an example of the prompt template used for inference in the Question Answering (QA) task. In the figure, the section highlighted in yellow is included only when the few-shot prompting option is enabled.

\paragraph{Dialogue Summarization.}
Figure~\ref{fig:prompt-dialogue-summarization} presents an example of the prompt template used for inference in the Dialogue Summarization task. Similar to the QA setting, the yellow-highlighted section is included only when the few-shot prompting option is applied.

\paragraph{Topic Classification.}
Figure~\ref{fig:prompt-topic-classification} illustrates an example of the prompt used for the Topic Classification task. This prompt uses one-shot setting and instructs the model to provide an explanation for its predicted label.

\paragraph{Machine-Generated Conversation.}
Figure~\ref{fig:prompt-machine-generated} illustrates an example of the prompt template used for inference in the Machine-Generated Conversation task. This prompt is used to perform inference with the GPT-4o model, enabling it to generate dialogue text that adheres to the specifications defined in the prompt.

\begin{figure*}[h]
\centering
\begin{tcolorbox}[
  enhanced,
  colback=green!5!white,
  colframe=green!75!black,
  title=Prompt example for Question Answering,
  width=\textwidth
]
Answer the question provided according to the context in the dialog. Just write the answer in the form of A, B, C, D, or E based on the appropriate option, without including the sentences accompanying the options. If the question does not match the context of the conversation or if none of the answer choices are correct, choose option "\textcolor{red}{<There is no correct answer>}".

\tcblower

\colorbox{yellow!15}{
\parbox{0.97\textwidth}{
Below are \textcolor{red}{<NUMBER OF SHOT>} example QA pairs from the example dialog.\\[0.5em]

\#\#\# EXAMPLE DIALOG\\
\textcolor{red}{<DIALOG EXAMPLE>}\\[0.5em]

\#\#\# EXAMPLE \textcolor{red}{<i>}\\
QUESTION: \textcolor{red}{<QUESTION EXAMPLE i>}\\
OPTIONS:\\
A. \textcolor{red}{<OPTION A>}\\
B. \textcolor{red}{<OPTION B>}\\
C. \textcolor{red}{<OPTION C>}\\
D. \textcolor{red}{<OPTION D>}\\
E. \textcolor{red}{<OPTION E>}\\
ANSWER: \textcolor{red}{<ANSWER EXAMPLE i>}\\
\\
Now answer the question for the dialog below.\\
}}
\\[1em]

\#\#\# DIALOG\\
\textcolor{red}{<DIALOG>}\\

\#\#\# QUESTION\\
\textcolor{red}{<QUESTION>}\\

\#\#\# OPTIONS\\
A. \textcolor{red}{<OPTION A>}\\
B. \textcolor{red}{<OPTION B>}\\
C. \textcolor{red}{<OPTION C>}\\
D. \textcolor{red}{<OPTION D>}\\
E. \textcolor{red}{<OPTION E>}\\

\#\#\# OUTPUT FORMAT\\
Return a JSON response in the following format:\\
\textcolor{red}{<OUTPUT FORMAT>}\\

\#\#\# OUTPUT
\end{tcolorbox}

\caption{Prompt template for the Question Answering task (Multiple Choice subset). The highlighted middle section corresponds to an optional few-shot component and is included only in the few-shot prompting setting.}
\label{fig:prompt-question-answering}
\end{figure*}

\begin{figure*}[h]
\centering
\begin{tcolorbox}[
  enhanced,
  colback=green!5!white,
  colframe=green!75!black,
  title=Prompt example for Dialogue Summarization,
  width=\textwidth
]
\# INSTRUCTIONS\\
Your primary task is to write a concise summary of a given dialogue. The summary should be focused on the provided topic and adhere strictly to the rules and output format specified below:\\
- The summary should be written in {target-lang}. Once again, write the summary in {target-lang}.\\
- The summary must be a single paragraph.\\
- The paragraph must be 3 to 5 sentences.\\
- The summary must be strictly factually correct, drawing all information directly from the dialogue. Do not introduce any external information or make assumptions not supported by the text.\\
- Prior to composing the summary, provide a brief analysis of the dialogue. This analysis should identify the key points selected for the summary and offer a justification for their inclusion.\\

\tcblower

\colorbox{yellow!15}{
\parbox{0.97\textwidth}{
\# EXAMPLES \\
\#\# DIALOGUE EXAMPLE\\
Topic: \textcolor{red}{<TOPIC EXAMPLE>}\\
Dialogue:\\
\textcolor{red}{<DIALOGUE EXAMPLE>}\\
\\
\#\# SUMMARY EXAMPLE \textcolor{red}{i}\\
\{\\
"summary": \textcolor{red}{SUMMARY i}\\
\}\\
\\
Now, please summarize the dialogue below.\\
}}
\\[1em]

\# DIALOGUE\\
Topic: \textcolor{red}{<TOPIC>}\\
Dialogue:\\
\textcolor{red}{<DIALOGUE>}\\
\\
\# OUTPUT FORMAT\\
Return a JSON response in the following format:\\
\textcolor{red}{<OUTPUT FORMAT>}\\

\# OUTPUT
\end{tcolorbox}

\caption{Prompt template for the Dialogue Summarization task. The highlighted middle section corresponds to an optional few-shot component and is included only in the few-shot prompting setting.}
\label{fig:prompt-dialogue-summarization}
\end{figure*}

\begin{figure*}[h]
\centering
\begin{tcolorbox}[
  enhanced,
  colback=green!5!white,
  colframe=green!75!black,
  title=Prompt example for Topic Classification,
  width=\textwidth
]
\# INSTRUCTIONS\\
Your primary task is to classify the provided code-switched/code-mixed dialogue into one of the following categories:\\
- Entertainment\\
- Science/Technology\\
- Social/Culture\\
- Education\\
- Daily Life\\
\\
Before classifying, please provide a brief analysis of the dialogue. This analysis should identify the key points that support your classification.\\
\\
\# DIALOGUE-TOPIC PAIR EXAMPLE\\
Dialogue:\\
\textcolor{red}{<DIALOG EXAMPLE>}\\
\\
\{ \\
"category": \textcolor{red}{<CATEGORY EXAMPLE>}\\
\} \\
\\
Now, please classify the dialogue below.\\
\\
\# DIALOGUE\\
Dialogue:\\
\textcolor{red}{<DIALOG>}\\
\\
\# OUTPUT FORMAT\\
Return a JSON response in the following format:\\
\{ \\
"explanation": "This section contains a brief analysis of the dialogue that supports the classification.",\\
"category": "This section contains exactly one category selected from the possible options. Output only the category name, without any additional text."\\
\} \\
\\
\# OUTPUT
\end{tcolorbox}

\caption{Prompt template for the Topic Classification.}
\label{fig:prompt-topic-classification}
\end{figure*}

\begin{figure*}[h]
\centering
\begin{tcolorbox}[
  enhanced,
  colback=green!5!white,
  colframe=green!75!black,
  title=Prompt example for Machine Generated--Conversational Text,
  width=\textwidth
]
Write a dialogue about \textcolor{red}{<TOPIC>} that uses code-switching and fulfills the following conditions:\\  
- The dialogue involves \textcolor{red}{<NUMBER OF PERSON>} people.  \\
- Use \textcolor{red}{<SPEAKER'S LABEL>} to label the speakers. \\ 
- The language used in the conversation should be code-switched between \textcolor{red}{<LANGUAGE>}. Make sure that each language is present in the conversation.\\
- The conversation is informal, so the use of slang, casual expressions, and relaxed punctuation, grammar, or capitalization is allowed.\\
- The dialogue should consist of 50 to 150 utterances (about 15 minutes conversation).  \\
- Emojis may be used in the dialogue.  \\
- The dialogue takes place in an online setting (e.g., group chat, direct messages).  \\
- Replies to specific messages are allowed and can be included.\\
- Ensure the conversation flows naturally.\\
\\
Use the following format for the dialogue: \\
\#1 A: {{insert text}}\\
\#2 B: {{insert text}}\\
\#3 C [reply to \#1]: {{insert text}}\\
\#4 A: {{insert text}}\\
\#5 D: {{insert text}}\\
\#6 A [reply to \#5]: {{insert text}}\\
\\
Write the dialogue here:
\end{tcolorbox}

\caption{Prompt template for the Machine Generated--Conversational Text.}
\label{fig:prompt-machine-generated}
\end{figure*}

\section{Hyperparameter Settings}
\label{sec:appendix_hyperparameter}

This section documents the hyperparameter configurations adopted for each model across the three downstream tasks. A complete summary of the settings is provided in Table~\ref{tab:appendix_hyperparameter_settings}. Given the substantial computational costs, all evaluations were limited to a single run. Because our core novelty lies in the dataset itself, our objective is not to provide an absolute model leaderboard, but rather to establish baseline performance across a diverse range of models and configurations.

\begin{table}[!th]
\centering
\resizebox{0.49\textwidth}{!}{
\begin{tabular}{@{}lccccc@{}}
\toprule
\textbf{Model} & 
    \textbf{Temperature} &
    \textbf{Top-P} &
    \textbf{Top-K} &
    \textbf{Min-P} &
    \textbf{Max-Tokens} \\
\midrule
\textbf{\textsc{Question Answering}} \\
\midrule
    Qwen2.5-3B-Instruct & 0.7 & 0.8 & 20 & 0 & 512 \\
    Qwen2.5-7B-Instruct & 0.7 & 0.8 & 20 & 0 & 512 \\
    Qwen3-4B \\
    $\quad$ Think & 0.6 & 0.95 & 0.20 & 0 & 8192 \\ 
    $\quad$ No-Think & 0.7 & 0.8 & 20 & 0 & 8192 \\ 
    Qwen3-8B \\
    $\quad$ Think & 0.6 & 0.95 & 0.20 & 0 & 8192 \\ 
    $\quad$ No-Think & 0.7 & 0.8 & 20 & 0 & 8192 \\ 
    Aya23-8B & 0.3 & 1.0 & 1 & 0 & 512 \\
    Gemma2-9B-Instruct & 0.1 & 0.9 & 5 & 0 & 512 \\
    Gemma3-4B-Instruct & 0.2 & 0.9 & 10 & 0 & 512 \\
    Sailor2-8B & 0.3 & 0.9 & 30 & 0 & 512 \\
    Sahabat-AI-Gemma & 0.1 & 0.9 & 5 & 0 & 512 \\
    SILMA-9B-Instruct & 0.3 & 0.9 & 10 & 0 & 512 \\
    ALLAM-7B-Instruct & 0.4 & 0.9 & 20 & 0 & 512 \\
\midrule
\multicolumn{6}{l}{\textbf{\textsc{Text Summarization \& Topic Classification}}} \\
\midrule
    Qwen2.5-3B-Instruct & 0.4 & 0.8 & 20 & 0 & 2048 \\
    Qwen2.5-7B-Instruct & 0.4 & 0.8 & 20 & 0 & 2048 \\
    Qwen3-4B \\
    $\quad$ Think    & 0.6 & 0.95 & 20 & 0 & 8192 \\
    $\quad$ No-Think & 0.7 & 0.8  & 20 & 0 & 2048 \\
    Qwen3-8B \\
    $\quad$ Think    & 0.6 & 0.95 & 20 & 0 & 8192 \\
    $\quad$ No-Think & 0.7 & 0.8  & 20 & 0 & 2048 \\
    Aya23-8B & 0.3 & 0.8 & 20 & 0 & 2048 \\
    Gemma2-9B-Instruct & 0.4 & 0.8 & 20 & 0 & 2048 \\
    Gemma3-4B-Instruct & 0.4 & 0.8 & 20 & 0 & 2048 \\
    Sailor2-8B-Chat & 0.4 & 0.8 & 20 & 0 & 2048 \\
    Sahabat-AI-Gemma & 0.4 & 0.8 & 20 & 0 & 2048 \\
    SILMA-9B-Instruct & 0.4 & 0.8 & 20 & 0 & 2048 \\
    ALLAM-7B-Instruct & 0.6 & 0.95 & 50 & 0 & 2048 \\
\bottomrule
\end{tabular}
}
\caption{Hyperparameter settings used for each model across all tasks.}
\label{tab:appendix_hyperparameter_settings}
\end{table}

\section{Qualitative Analysis of QA Topic}
\label{sec:appendix_qualitative_analysis}

Based on the ablation results presented in Table~\ref{tab:ablation_experiment_result}, several language configurations consistently exhibit substantially lower performance, often falling below random-chance level (<50\%), particularly Indonesian-English (ID-EN) and Javanese-Indonesian-English (JV-ID-EN). We hypothesize that this degradation primarily stems from the non-parallel nature of our dataset construction. Unlike parallel multilingual benchmarks, each language configuration in our dataset contains distinct forms of local knowledge embedded within the dialogue context, alongside varying conversational topics and culturally grounded references across language pairs.

\begin{table*}[!th]
\renewcommand{\arraystretch}{1.4}
\centering
\resizebox{\textwidth}{!}{
\begin{tabular}{lcclc|cclc}
\toprule
&\multicolumn{4}{c|}{\textsc{\textbf{ID-EN}}} & \multicolumn{4}{c}{\textsc{\textbf{JV-ID-EN}}} \\
\cmidrule(lr){2-5} \cmidrule(lr){6-9}
 & Avg. (\%)
 & ID
 & Topic
 & Too Specific (?)
 & Avg. (\%)
 & ID
 & Topic
 & Too Specific (?)
 \\
\midrule
\textbf{Top-10} \\
\midrule
& 47.78 & 19 & All about BlackPink & \xmark & 60.00 & 22 & Experiences of being admitted to medical school & \cmark \\ 
& 47.78 & 69 & Talking about the latest chatbots/LLMs & \xmark & 60.00 & 88 & Discussions on patents and copyright protection & \xmark \\
& 45.56 & 17 & All about bioinformatics & \xmark & 58.89 & 25 & The lack of digital literacy interest among Indonesian youth & \xmark \\
& 44.44 & 16 & All about anime, favorites or recently watched & \xmark & 55.56 & 90 & Tourism villages and local community development & \xmark \\
& 44.44 & 75 & Discussion on ISAE-SUPAERO’s master’s programs & \cmark & 54.44 & 13 & Experiences of participating in debate competitions during high school & \cmark \\
& 43.33 & 20 & Recently played games & \cmark & 54.44 & 46 & Interest in learning foreign languages & \xmark \\
& 42.22 & 2 & Current developments in electric vehicle technology & \xmark & 53.33 & 24 & Perspectives on the legalization of cannabis for medical purposes & \xmark \\
& 41.11 & 76 & Talking about various renewable energy sources & \xmark & 53.33 & 53 & Being a morning person versus a night owl & \xmark \\
& 40.00 & 68 & Discussing the Israel-Hamas ceasefire & \xmark & 52.22 & 2 & Opinions on Manchester United & \xmark \\
& 40.00 & 81 & Discussing Trump’s new policies & \xmark & 52.22 & 15 & Experiences of studying in the US & \xmark \\
\midrule
\textbf{Bottom-10} \\
\midrule
& 10.00 & 92 & Discussing various music platforms & \xmark & 17.78 & 40 & Perspectives on the COC event by Ruangguru & \cmark \\
& 13.33 & 99 & Public transportation in Bandung & \cmark & 20.00 & 61 & Experiences of participating in “mlaku-mlaku” activities & \cmark \\
& 14.44 & 95 & Complaints about Edunex & \cmark & 24.44 & 8 & Perspectives on the inauguration of Indonesia’s new president in 2024 & \xmark \\
& 15.56 & 79 & Pros and cons of ITB’s fast-track program & \cmark & 24.44 & 34 & Weather predictions for Indonesian cities based recent days & \cmark \\
& 16.67 & 15 & Current meta heroes in Mobile Legends & \cmark & 24.44 & 76 & Experiences of falling before a proposal seminar presentation & \cmark \\
& 16.67 & 18 & Thoughts on Apple’s M4 chip & \xmark & 26.67 & 70 & Discussions about proposal seminars (seminar proposal / sempro) & \cmark \\
& 16.67 & 61 & Thoughts on Indonesia joining BRICS & \xmark & 27.78 & 3 & Experiences related to local culinary traditions & \cmark \\
& 17.78 & 22 & Recent activities & \cmark & 27.78 & 35 & Favorite music genres and musicians & \cmark \\
& 17.78 & 63 & Favorite Indonesian dishes & \cmark & 27.78 & 56 & Evidence-Based Case Report examination for medical students & \xmark \\
& 17.78 & 48 & Thoughts on Academy of Champions (CoC for high schoolers) & \cmark & 27.78 & 71 & Research experiences in Kalimantan & \cmark \\
\bottomrule
\renewcommand{\arraystretch}{1}
\end{tabular}
}
\caption{Qualitative analysis of the Top-10 and Bottom-10 dialogues in the ID-EN and JV-ID-EN configurations based on the average performance of all evaluated LLMs. \textbf{Avg.} denotes the average model performance score, \textbf{ID} refers to the dialogue identifier, \textbf{Topic} describes the dialogue content, and \textbf{Too Specific (?)} indicates whether annotators considered the dialogue excessively specific due to local knowledge, personal experiences, or temporally grounded contexts.}
\label{tab:qualitative_analysis}
\vspace{-1em}
\end{table*}

To further investigate this phenomenon, Table~\ref{tab:qualitative_analysis} presents a qualitative analysis of the two lowest-performing language configurations. For each language pair, we selected the top-10 and bottom-10 dialogues based on the average performance of all evaluated LLMs. Human annotators were then asked to assess whether the dialogue topics were excessively specific, such as containing strong local knowledge dependencies, references tied to the original annotator’s personal experiences, or temporally constrained contexts that may not generalize across linguistic settings.

The analysis reveals a clear trend. In the ID-EN configuration, 7 out of the 10 lowest-performing dialogues were categorized as overly specific. A similar pattern was observed in JV-ID-EN, where 8 out of the 10 bottom-ranked dialogues contained highly specific contextual content. In contrast, among the top-10 dialogues with the highest average model performance, only 2 dialogues in both ID-EN and JV-ID-EN were considered excessively specific. These findings provide supporting evidence for our hypothesis that the non-parallel structure of the dataset introduces substantial contextual variability across language configurations.

\section{Qualitative Analysis of Summarization}
\label{sec:qualitative_summarization}
 
Beyond the quantitative metrics, we examine specific failure patterns observed in the summarization outputs to better understand model behavior across languages and prompting conditions.
 
\subsection{Few-Shot Context Overflow}
 
A notable failure was observed for \textbf{Sailor2-8B} and \textbf{ALLAM-7B-Instruct} under few-shot settings. Both models have a maximum context length of 4,096 tokens, and when few-shot examples are included in the prompt, the total input length frequently exceeds this limit. As a result, the API returns an HTTP 400 error of the following form:
 
\begin{quote}
\small\ttfamily ``This model's maximum context length is 4096 tokens. However, you requested 2048 output tokens and your prompt contains at least 2049 input tokens, for a total of at least 4097 tokens.''
\end{quote}
 
This explains the dramatic BERTScore collapse observed for both models at 1-shot and 3-shot settings - the outputs are not degenerate summaries, but rather missing entirely due to failed API calls. This is a practical limitation worth noting when deploying smaller-context models in few-shot summarization pipelines.
 
\subsection{Script Inconsistency in Algerian Arabic}
 
For the AR-DZ-FR language pair, the expected output is \textit{Arabizi} - a transliteration of Algerian Arabic using Latin characters mixed with numerals. However, we observe that several models occasionally produce summaries in Arabic script instead, despite the input dialogue being written in Arabizi. This script mismatch directly penalises surface-level metrics such as ROUGE, since character-level overlap between Arabic script and Arabizi is effectively zero even when the semantic content is correct. This explains the notably higher BERTScore-F values relative to ROUGE on this language pair - the model captures meaning but renders it in the wrong script. More broadly, Arabizi lacks a standardised orthography and is not formally supported in most model training pipelines, making consistent script adherence difficult to enforce without explicit instruction.

\section{Quantitative Results}
\label{sec:quantitative_result}

This section provides a comprehensive overview of the quantitative findings obtained under all experimental configurations.

\subsection{Question Answering}
\label{sec:quantitative-results-question-answering}

For the Question Answering task, results are analyzed from two complementary perspectives: (1) the effect of reasoning versus non-reasoning settings, and (2) the impact of zero-shot versus one-shot prompting strategies.

\subsubsection{Reasoning vs. Non-Reasoning}
\label{sec:quantitative-results-question-answering-reasoning-non}

Table~\ref{tab:qa_reasoning_non_statistics} summarizes the performance comparison between reasoning-enabled and non-reasoning configurations.

\begin{table*}[!th]
\centering
\resizebox{\textwidth}{!}{
\begin{tabular}{@{}lcccccccccc@{}}
\toprule
& \multicolumn{2}{c}{\textsc{ID-EN}} & \multicolumn{2}{c}{\textsc{JV-ID-EN}} & \multicolumn{2}{c}{\textsc{SU-ID-EN}} & \multicolumn{2}{c}{\textsc{HA-EN}} & \multicolumn{2}{c}{\textsc{ar\_DZ-FR}}\\
\cmidrule(lr){2-3} \cmidrule(lr){4-5} \cmidrule(lr){6-7} \cmidrule(lr){8-9} \cmidrule(lr){10-11}
\textbf{model} & 
    \textbf{Reasoning} & 
    \textbf{Non-reasoning} & 
    \textbf{Reasoning} & 
    \textbf{Non-reasoning} & 
    \textbf{Reasoning} & 
    \textbf{Non-reasoning} & 
    \textbf{Reasoning} & 
    \textbf{Non-reasoning} & 
    \textbf{Reasoning} & 
    \textbf{Non-reasoning} \\
\midrule
\textbf{Global} \\
\midrule
    Qwen2.5-3B-Instruct & 11.51 & 5.85 & 25.85 & 18.98 & 86.66 & 76.16 & 43.93 & 43.94 & 91.91 & 25.25 \\
    Qwen2.5-7B-Instruct & 12.12 & 4.64 & 35.75 & 24.64 & 90.50 & 86.68 & 32.82 & 43.93 & 40.40 & 38.37 \\
    Qwen3-4B & 15.55 & 8.08 & 42.22 & 30.90 & 90.70 & 89.29 & 49.49 & 55.55 & 39.39 & 32.32 \\
    Qwen3-8B & 21.41 & 8.68 & 46.66 & 31.71 & 89.89 & 89.29 & 50.00 & 45.95 & 41.41 & 44.44 \\
    Aya23-8B & 25.65 & 8.28 & 31.51 & 23.03 & 86.46 & 80.00 & 29.79 & 28.78 & 28.28 & 14.14 \\
    Gemma2-9B-Instruct & 13.73 & 11.91 & 45.05 & 43.63 & 91.71 & 91.91 & 82.82 & 82.82 & 52.52 & 50.50 \\
    Gemma3-4B-Instruct & 25.05 & 23.43 & 41.01 & 36.76 & 91.31 & 90.70 & 71.21 & 69.19 & 36.36 & 39.39 \\
\midrule
\textbf{Regional} \\
\midrule
    Sailor2-8B & 35.55 & 27.67 & 37.17 & 40.40 & 86.66 & 86.46 & - & - & - & - \\
    Sahabat-AI-Gemma & 16.16 & 13.73 & 46.06 & 43.23 & 93.13 & 93.53 & - & - & - & - \\
    SILMA-9B-Instruct & - & - &  - & - & - & - & - & - & 0.0 & 0.0 \\
    ALLAM-7B-Instruct & - & - &  - & - & - & - & - & - & 39.39 & 32.32 \\
\bottomrule
\end{tabular}
}
\caption{Statistics (Acc. \%) per-language combination of Reasoning vs. non-Reasoning approach on Question Answering task}
\label{tab:qa_reasoning_non_statistics}
\end{table*}

\subsubsection{0-shot vs. 1-shot}
\label{sec:quantitative-results-question-answering-shot}

The comparative results for zero-shot and one-shot prompting setups are presented in Table~\ref{tab:qa_shot_statistics}.

\begin{table*}[!th]
\centering
\resizebox{\textwidth}{!}{
\begin{tabular}{@{}lcccccccccc@{}}
\toprule
& \multicolumn{2}{c}{\textsc{ID-EN}} & \multicolumn{2}{c}{\textsc{JV-ID-EN}} & \multicolumn{2}{c}{\textsc{SU-ID-EN}} & \multicolumn{2}{c}{\textsc{HA-EN}} & \multicolumn{2}{c}{\textsc{ar\_DZ-FR}}\\
\cmidrule(lr){2-3} \cmidrule(lr){4-5} \cmidrule(lr){6-7} \cmidrule(lr){8-9} \cmidrule(lr){10-11}
\textbf{model} & 
    \textbf{0-shot} & 
    \textbf{1-shot} & 
    \textbf{0-shot} & 
    \textbf{1-shot} & 
    \textbf{0-shot} & 
    \textbf{1-shot} & 
    \textbf{0-shot} & 
    \textbf{1-shot} & 
    \textbf{0-shot} & 
    \textbf{1-shot} \\
\midrule
\textbf{Global} \\
\midrule
    Qwen2.5-3B-Instruct & 5.85 & 4,64 & 18.08 & 19.19 & 76.16 & 72.32 & 43.93 & 42.42 & 25.25 & 25.25 \\
    Qwen2.5-7B-Instruct & 4.64 & 3.63 & 24.64 & 24.24 & 88.68 & 87.07 & 43.93 & 42.92 & 38.38 & 33.33 \\
    Qwen3-4B & 8.88 & 8.08 & 30.90 & 28.08 & 90.70 & 89.09 & 49.49 & 51.01 & 39.39 & 38.38 \\
    Qwen3-8B & 5.65 & 7.07 & 31.71 & 31.31 & 89.89 & 89.29 & 50.00 & 44.44 & 41.41 & 37.37 \\
    Aya23-8B & 8.28 & 13.93 & 23.03 & 28.08 & 80.0 & 84.24 & 28.78 & 26.76 & 14.14 & 13.13 \\
    Gemma2-9B-Instruct & 11.91 & 9.09 & 43.63 & 39.19 & 91.91 & 91.51 & 82.82 & 81.81 & 50.50 & 49.49 \\
    Gemma3-4B-Instruct & 23.43 & 20.40 & 36.76 & 33.93 & 90.70 & 90.50 & 69.19 & 62.62 & 39.39 & 38.38 \\
\midrule
\textbf{Regional} \\
\midrule
    Sailor2-8B & 27.67 & 29.89 & 40.40 & 39.19 & 86.46 & 85.45 & - & - & - & - \\
    Sahabat-AI-Gemma & 13.73 & 12.12 & 43.23 & 41.01 & 93.53 & 93.53 & - & - & - & - \\
    SILMA-9B-Instruct & - & - &  - & - & - & - & - & - & 0.0 & 0.0 \\
    ALLAM-7B-Instruct & - & - &  - & - & - & - & - & - & 32.32 & 35.35 \\
\bottomrule
\end{tabular}
}
\caption{Statistics (Acc. \%) per-language combination of 0-shot vs. 1-shot approach on Question Answering task}
\label{tab:qa_shot_statistics}
\end{table*}

\subsection{Text Summarization}
\label{sec:quantitative-results-summary}

For the Text Summarization task, we similarly examine two major dimensions: the role of reasoning mechanisms and the effect of prompt shot size.

\subsubsection{Reasoning vs. Non-Reasoning}
\label{sec:quantitative-results-summarys-reasoning-non}

Detailed results comparing reasoning and non-reasoning settings across different language combinations are reported in Table~\ref{tab:summary_reasoning_non_statistics_id-en}, Table~\ref{tab:summary_reasoning_non_statistics_jv-id-en}, Table~\ref{tab:summary_reasoning_non_statistics_su-id-en}, Table~\ref{tab:summary_reasoning_non_statistics_ha-en}, and Table~\ref{tab:summary_reasoning_non_statistics_ar-dz-fr}.

\begin{table*}[!th]
\centering
\renewcommand{\arraystretch}{1.25}
\resizebox{\textwidth}{!}{
\begin{tabular}{@{}lcccccccccc@{}}
\toprule
& & \multicolumn{9}{c}{\textsc{ID-EN}}\\
\cmidrule(lr){3-11}
\textbf{model} & \textbf{Reasoning} & 
    \textbf{ROUGE1} &
    \textbf{ROUGE2} &
    \textbf{ROUGE3} &
    \textbf{ROUGE4} &
    \textbf{METEOR} &
    \textbf{CHRF++} &
    \textbf{BERTScore-p} &
    \textbf{BERTScore-r} &
    \textbf{BERTScore-f} \\
\midrule
\textbf{Global} \\
\midrule
    Qwen2.5-3B-Instruct & \cmark & 0.292 & 0.078 & 0.027 & 0.010 & 0.216 & 32.022 & 0.903 & 0.896 & 0.895 \\
    Qwen2.5-3B-Instruct & \xmark & 0.343 & 0.084 & 0.029 & 0.010 & 0.263 & 36.575 & 0.896 & 0.902 & 0.894 \\
    Qwen2.5-7B-Instruct & \cmark & 0.337 & 0.102 & 0.042 & 0.017 & 0.227 & 32.942 & 0.910 & 0.899 & 0.899 \\
    Qwen2.5-7B-Instruct & \xmark & 0.328 & 0.090 & 0.033 & 0.013 & 0.274 & 36.315 & 0.878 & 0.905 & 0.886 \\
    Qwen3-4B & \cmark & 0.352 & 0.102 & 0.039 & 0.016 & 0.258 & 36.069 & 0.904 & 0.904 & 0.899 \\
    Qwen3-4B & \xmark & 0.361 & 0.102 & 0.038 & 0.014 & 0.276 & 37.676 & 0.898 & 0.905 & 0.897 \\
    Qwen3-8B & \cmark & 0.342 & 0.099 & 0.037 & 0.015 & 0.248 & 35.331 & 0.904 & 0.902 & 0.898 \\
    Qwen3-8B & \xmark & 0.242 & 0.070 & 0.027 & 0.011 & 0.206 & 26.973 & 0.867 & 0.903 & 0.880 \\
    Aya23-8B & \cmark & 0.203 & 0.053 & 0.020 & 0.008 & 0.170 & 27.000 & 0.894 & 0.890 & 0.888 \\
    Aya23-8B & \xmark & 0.347 & 0.098 & 0.037 & 0.015 & 0.275 & 38.158 & 0.898 & 0.904 & 0.896 \\
    Gemma2-9B-Instruct & \cmark & 0.348 & 0.114 & 0.049 & 0.024 & 0.253 & 35.622 & 0.909 & 0.905 & 0.902 \\
    Gemma2-9B-Instruct & \xmark & 0.396 & 0.126 & 0.054 & 0.025 & 0.303 & 40.285 & 0.905 & 0.910 & 0.903 \\
    Gemma3-4B-Instruct & \cmark & 0.365 & 0.106 & 0.040 & 0.017 & 0.285 & 38.663 & 0.897 & 0.904 & 0.897 \\
    Gemma3-4B-Instruct & \xmark & 0.287 & 0.079 & 0.030 & 0.012 & 0.270 & 36.126 & 0.847 & 0.905 & 0.872 \\
\midrule
\textbf{Regional} \\
\midrule
    Sailor2-8B & \cmark & 0.243 & 0.059 & 0.020 & 0.007 & 0.169 & 25.686 & 0.749 & 0.750 & 0.745 \\
    Sailor2-8B & \xmark & 0.278 & 0.067 & 0.023 & 0.007 & 0.203 & 29.451 & 0.732 & 0.746 & 0.736 \\
    Sahabat-AI-Gemma & \cmark & 0.349 & 0.110 & 0.045 & 0.022 & 0.255 & 35.563 & 0.909 & 0.903 & 0.901 \\
    Sahabat-AI-Gemma & \xmark & 0.370 & 0.122 & 0.048 & 0.021 & 0.293 & 39.021 & 0.901 & 0.907 & 0.899 \\
\bottomrule
\end{tabular}
}
\caption{Statistics for the remaining evaluation metrics (ROUGE-1, ROUGE-2, ROUGE-3, ROUGE-4, METEOR, CHRF++, and BERTScore-P/R/F) on the Indonesian-English (ID-EN) language pair, comparing reasoning and non-reasoning approaches for the Text Summarization task.}
\label{tab:summary_reasoning_non_statistics_id-en}
\renewcommand{\arraystretch}{1}
\end{table*}

\begin{table*}[!th]
\centering
\renewcommand{\arraystretch}{1.25}
\resizebox{\textwidth}{!}{
\begin{tabular}{@{}lcccccccccc@{}}
\toprule
& & \multicolumn{9}{c}{\textsc{JV-ID-EN}}\\
\cmidrule(lr){3-11}
\textbf{model} & \textbf{Reasoning} & 
    \textbf{ROUGE1} &
    \textbf{ROUGE2} &
    \textbf{ROUGE3} &
    \textbf{ROUGE4} &
    \textbf{METEOR} &
    \textbf{CHRF++} &
    \textbf{BERTScore-p} &
    \textbf{BERTScore-r} &
    \textbf{BERTScore-f} \\
\midrule
\textbf{Global} \\
\midrule
    Qwen2.5-3B-Instruct & \cmark & 0.298 & 0.071 & 0.023 & 0.008 & 0.189 & 30.166 & 0.904 & 0.884 & 0.892 \\
    Qwen2.5-3B-Instruct & \xmark & 0.345 & 0.084 & 0.030 & 0.012 & 0.251 & 37.283 & 0.895 & 0.894 & 0.893 \\
    Qwen2.5-7B-Instruct & \cmark & 0.312 & 0.088 & 0.032 & 0.011 & 0.187 & 29.252 & 0.910 & 0.886 & 0.896 \\
    Qwen2.5-7B-Instruct & \xmark & 0.311 & 0.079 & 0.029 & 0.010 & 0.247 & 36.214 & 0.882 & 0.892 & 0.885 \\
    Qwen3-4B & \cmark & 0.336 & 0.082 & 0.030 & 0.010 & 0.221 & 34.275 & 0.903 & 0.890 & 0.895 \\
    Qwen3-4B & \xmark & 0.353 & 0.089 & 0.033 & 0.013 & 0.249 & 37.348 & 0.900 & 0.895 & 0.896 \\
    Qwen3-8B & \cmark & 0.327 & 0.084 & 0.027 & 0.007 & 0.216 & 33.971 & 0.902 & 0.891 & 0.895 \\
    Qwen3-8B & \xmark & 0.266 & 0.071 & 0.025 & 0.009 & 0.199 & 28.539 & 0.872 & 0.894 & 0.881 \\
    Aya23-8B & \cmark & 0.200 & 0.047 & 0.016 & 0.005 & 0.143 & 24.163 & 0.891 & 0.878 & 0.883 \\
    Aya23-8B & \xmark & 0.349 & 0.092 & 0.034 & 0.012 & 0.252 & 38.227 & 0.897 & 0.892 & 0.894 \\
    Gemma2-9B-Instruct & \cmark & 0.319 & 0.086 & 0.030 & 0.011 & 0.206 & 31.614 & 0.906 & 0.889 & 0.896 \\
    Gemma2-9B-Instruct & \xmark & 0.369 & 0.108 & 0.043 & 0.017 & 0.270 & 38.801 & 0.901 & 0.898 & 0.898 \\
    Gemma3-4B-Instruct & \cmark & 0.349 & 0.085 & 0.029 & 0.012 & 0.248 & 37.752 & 0.895 & 0.894 & 0.893 \\
    Gemma3-4B-Instruct & \xmark & 0.238 & 0.059 & 0.020 & 0.007 & 0.228 & 32.902 & 0.826 & 0.895 & 0.857 \\
\midrule
\textbf{Regional} \\
\midrule
    Sailor2-8B & \cmark & 0.207 & 0.039 & 0.011 & 0.002 & 0.128 & 22.042 & 0.700 & 0.692 & 0.695 \\
    Sailor2-8B & \xmark & 0.225 & 0.047 & 0.015 & 0.004 & 0.156 & 25.473 & 0.658 & 0.664 & 0.660 \\
    Sahabat-AI-Gemma & \cmark & 0.330 & 0.098 & 0.041 & 0.018 & 0.213 & 31.905 & 0.907 & 0.887 & 0.895 \\
    Sahabat-AI-Gemma & \xmark & 0.354 & 0.107 & 0.044 & 0.018 & 0.256 & 37.021 & 0.900 & 0.893 & 0.895 \\
\bottomrule
\end{tabular}
}
\caption{Statistics for the remaining evaluation metrics (ROUGE-1, ROUGE-2, ROUGE-3, ROUGE-4, METEOR, CHRF++, and BERTScore-P/R/F) on the Javanese-Indonesian-English (JV-ID-EN) language pair, comparing reasoning and non-reasoning approaches for the Text Summarization task.}
\label{tab:summary_reasoning_non_statistics_jv-id-en}
\renewcommand{\arraystretch}{1}
\end{table*}

\begin{table*}[!th]
\centering
\renewcommand{\arraystretch}{1.25}
\resizebox{\textwidth}{!}{
\begin{tabular}{@{}lcccccccccc@{}}
\toprule
& & \multicolumn{9}{c}{\textsc{SU-ID-EN}}\\
\cmidrule(lr){3-11}
\textbf{model} & \textbf{Reasoning} & 
    \textbf{ROUGE1} &
    \textbf{ROUGE2} &
    \textbf{ROUGE3} &
    \textbf{ROUGE4} &
    \textbf{METEOR} &
    \textbf{CHRF++} &
    \textbf{BERTScore-p} &
    \textbf{BERTScore-r} &
    \textbf{BERTScore-f} \\
\midrule
\textbf{Global} \\
\midrule
    Qwen2.5-3B-Instruct & \cmark & 0.346 & 0.109 & 0.043 & 0.017 & 0.219 & 32.329 & 0.914 & 0.890 & 0.899 \\
    Qwen2.5-3B-Instruct & \xmark & 0.394 & 0.119 & 0.046 & 0.018 & 0.278 & 39.479 & 0.908 & 0.898 & 0.901 \\
    Qwen2.5-7B-Instruct & \cmark & 0.348 & 0.119 & 0.051 & 0.024 & 0.203 & 30.222 & 0.920 & 0.889 & 0.902 \\
    Qwen2.5-7B-Instruct & \xmark & 0.372 & 0.121 & 0.047 & 0.019 & 0.276 & 38.705 & 0.897 & 0.898 & 0.895 \\
    Qwen3-4B & \cmark & 0.397 & 0.126 & 0.050 & 0.022 & 0.253 & 36.549 & 0.918 & 0.897 & 0.905 \\
    Qwen3-4B & \xmark & 0.420 & 0.137 & 0.057 & 0.026 & 0.280 & 39.511 & 0.915 & 0.900 & 0.905 \\
    Qwen3-8B & \cmark & 0.382 & 0.123 & 0.049 & 0.022 & 0.238 & 34.731 & 0.917 & 0.895 & 0.903 \\
    Qwen3-8B & \xmark & 0.324 & 0.107 & 0.043 & 0.018 & 0.234 & 31.775 & 0.893 & 0.898 & 0.893 \\
    Aya23-8B & \cmark & 0.264 & 0.082 & 0.035 & 0.015 & 0.171 & 27.423 & 0.907 & 0.883 & 0.892 \\
    Aya23-8B & \xmark & 0.404 & 0.134 & 0.055 & 0.023 & 0.281 & 40.073 & 0.911 & 0.898 & 0.902 \\
    Gemma2-9B-Instruct & \cmark & 0.391 & 0.141 & 0.063 & 0.028 & 0.252 & 34.637 & 0.922 & 0.895 & 0.906 \\
    Gemma2-9B-Instruct & \xmark & 0.441 & 0.161 & 0.074 & 0.037 & 0.309 & 41.759 & 0.917 & 0.904 & 0.908 \\
    Gemma3-4B-Instruct & \cmark & 0.422 & 0.144 & 0.059 & 0.028 & 0.299 & 40.948 & 0.914 & 0.903 & 0.906 \\
    Gemma3-4B-Instruct & \xmark & 0.328 & 0.110 & 0.046 & 0.021 & 0.294 & 38.784 & 0.852 & 0.905 & 0.875 \\
\midrule
\textbf{Regional} \\
\midrule
    Sailor2-8B & \cmark & 0.210 & 0.062 & 0.021 & 0.008 & 0.128 & 19.015 & 0.592 & 0.581 & 0.585 \\
    Sailor2-8B & \xmark & 0.258 & 0.075 & 0.026 & 0.010 & 0.172 & 24.393 & 0.588 & 0.590 & 0.588 \\
    Sahabat-AI-Gemma & \cmark & 0.400 & 0.153 & 0.069 & 0.034 & 0.257 & 35.298 & 0.919 & 0.896 & 0.905 \\
    Sahabat-AI-Gemma & \xmark & 0.448 & 0.163 & 0.073 & 0.037 & 0.309 & 41.698 & 0.917 & 0.902 & 0.907 \\
\bottomrule
\end{tabular}
}
\caption{Statistics for the remaining evaluation metrics (ROUGE-1, ROUGE-2, ROUGE-3, ROUGE-4, METEOR, CHRF++, and BERTScore-P/R/F) on the Sundanese-Indonesian-English (SU-ID-EN) language pair, comparing reasoning and non-reasoning approaches for the Text Summarization task.}
\label{tab:summary_reasoning_non_statistics_su-id-en}
\renewcommand{\arraystretch}{1}
\end{table*}

\begin{table*}[!th]
\centering
\renewcommand{\arraystretch}{1.25}
\resizebox{\textwidth}{!}{
\begin{tabular}{@{}lcccccccccc@{}}
\toprule
& & \multicolumn{9}{c}{\textsc{HA-EN}}\\
\cmidrule(lr){3-11}
\textbf{model} & \textbf{Reasoning} & 
    \textbf{ROUGE1} &
    \textbf{ROUGE2} &
    \textbf{ROUGE3} &
    \textbf{ROUGE4} &
    \textbf{METEOR} &
    \textbf{CHRF++} &
    \textbf{BERTScore-p} &
    \textbf{BERTScore-r} &
    \textbf{BERTScore-f} \\
\midrule
\textbf{Global} \\
\midrule
    Qwen2.5-3B-Instruct & \cmark & 0.060 & 0.007 & 0.002 & 0.000 & 0.059 & 9.222 & 0.757 & 0.788 & 0.770 \\
    Qwen2.5-3B-Instruct & \xmark & 0.063 & 0.009 & 0.002 & 0.001 & 0.071 & 9.568 & 0.758 & 0.802 & 0.777 \\
    Qwen2.5-7B-Instruct & \cmark & 0.216 & 0.032 & 0.006 & 0.001 & 0.130 & 19.915 & 0.842 & 0.836 & 0.837 \\
    Qwen2.5-7B-Instruct & \xmark & 0.176 & 0.029 & 0.009 & 0.003 & 0.131 & 18.438 & 0.814 & 0.835 & 0.822 \\
    Qwen3-4B & \cmark & 0.062 & 0.007 & 0.001 & 0.000 & 0.047 & 7.423 & 0.758 & 0.776 & 0.765 \\
    Qwen3-4B & \xmark & 0.077 & 0.011 & 0.003 & 0.002 & 0.066 & 9.369 & 0.764 & 0.789 & 0.774 \\
    Qwen3-8B & \cmark & 0.038 & 0.005 & 0.001 & 0.000 & 0.031 & 4.755 & 0.756 & 0.764 & 0.758 \\
    Qwen3-8B & \xmark & 0.017 & 0.004 & 0.001 & 0.000 & 0.024 & 3.206 & 0.752 & 0.772 & 0.759 \\
    Aya23-8B & \cmark & 0.127 & 0.021 & 0.006 & 0.001 & 0.111 & 17.010 & 0.817 & 0.824 & 0.818 \\
    Aya23-8B & \xmark & 0.097 & 0.017 & 0.005 & 0.002 & 0.097 & 14.827 & 0.792 & 0.820 & 0.803 \\
    Gemma2-9B-Instruct & \cmark & 0.350 & 0.094 & 0.033 & 0.012 & 0.212 & 30.277 & 0.891 & 0.871 & 0.879 \\
    Gemma2-9B-Instruct & \xmark & 0.360 & 0.098 & 0.036 & 0.015 & 0.243 & 33.598 & 0.884 & 0.878 & 0.879 \\
    Gemma3-4B-Instruct & \cmark & 0.261 & 0.049 & 0.013 & 0.003 & 0.188 & 26.318 & 0.841 & 0.859 & 0.848 \\
    Gemma3-4B-Instruct & \xmark & 0.160 & 0.032 & 0.009 & 0.003 & 0.150 & 19.467 & 0.805 & 0.856 & 0.827 \\
\bottomrule
\end{tabular}
}
\caption{Statistics for the remaining evaluation metrics (ROUGE-1, ROUGE-2, ROUGE-3, ROUGE-4, METEOR, CHRF++, and BERTScore-P/R/F) on the Hausa-English (HA-EN) language pair, comparing reasoning and non-reasoning approaches for the Text Summarization task.}
\label{tab:summary_reasoning_non_statistics_ha-en}
\renewcommand{\arraystretch}{1}
\end{table*}

\begin{table*}[!th]
\centering
\renewcommand{\arraystretch}{1.25}
\resizebox{\textwidth}{!}{
\begin{tabular}{@{}lcccccccccc@{}}
\toprule
& & \multicolumn{9}{c}{\textsc{ar\_DZ-FR}}\\
\cmidrule(lr){3-11}
\textbf{model} & \textbf{Reasoning} & 
    \textbf{ROUGE1} &
    \textbf{ROUGE2} &
    \textbf{ROUGE3} &
    \textbf{ROUGE4} &
    \textbf{METEOR} &
    \textbf{CHRF++} &
    \textbf{BERTScore-p} &
    \textbf{BERTScore-r} &
    \textbf{BERTScore-f} \\
\midrule
\textbf{Global} \\
\midrule
    Qwen2.5-3B-Instruct & \cmark & 0.018 & 0.001 & 0.000 & 0.000 & 0.039 & 3.756 & 0.806 & 0.781 & 0.792 \\
    Qwen2.5-3B-Instruct & \xmark & 0.006 & 0.001 & 0.001 & 0.001 & 0.032 & 1.184 & 0.807 & 0.766 & 0.784 \\
    Qwen2.5-7B-Instruct & \cmark & 0.149 & 0.022 & 0.006 & 0.002 & 0.109 & 19.202 & 0.850 & 0.845 & 0.847 \\
    Qwen2.5-7B-Instruct & \xmark & 0.083 & 0.015 & 0.005 & 0.002 & 0.091 & 11.835 & 0.797 & 0.830 & 0.812 \\
    Qwen3-4B & \cmark & 0.093 & 0.015 & 0.004 & 0.001 & 0.076 & 12.491 & 0.815 & 0.813 & 0.812 \\
    Qwen3-4B & \xmark & 0.087 & 0.019 & 0.006 & 0.002 & 0.078 & 11.808 & 0.794 & 0.814 & 0.802 \\
    Qwen3-8B & \cmark & 0.086 & 0.013 & 0.002 & 0.000 & 0.076 & 12.815 & 0.802 & 0.819 & 0.809 \\
    Qwen3-8B & \xmark & 0.025 & 0.005 & 0.002 & 0.000 & 0.033 & 4.445 & 0.773 & 0.805 & 0.787 \\
    Aya23-8B & \cmark & 0.054 & 0.009 & 0.002 & 0.001 & 0.073 & 14.529 & 0.820 & 0.783 & 0.800 \\
    Aya23-8B & \xmark & 0.042 & 0.010 & 0.003 & 0.001 & 0.068 & 9.450 & 0.798 & 0.779 & 0.786 \\
    Gemma2-9B-Instruct & \cmark & 0.138 & 0.027 & 0.009 & 0.003 & 0.100 & 16.490 & 0.838 & 0.822 & 0.829 \\
    Gemma2-9B-Instruct & \xmark & 0.108 & 0.019 & 0.007 & 0.002 & 0.107 & 16.018 & 0.810 & 0.817 & 0.812 \\
    Gemma3-4B-Instruct & \cmark & 0.033 & 0.005 & 0.001 & 0.000 & 0.048 & 4.412 & 0.787 & 0.787 & 0.785 \\
    Gemma3-4B-Instruct & \xmark & 0.020 & 0.004 & 0.002 & 0.001 & 0.032 & 2.463 & 0.758 & 0.776 & 0.764 \\
\midrule
\textbf{Regional} \\
\midrule
    SILMA-9B-Instruct & \cmark & 0.025 & 0.007 & 0.003 & 0.001 & 0.030 & 4.326 & 0.798 & 0.765 & 0.779 \\
    SILMA-9B-Instruct & \xmark & 0.010 & 0.002 & 0.000 & 0.000 & 0.010 & 1.230 & 0.715 & 0.756 & 0.733 \\
    ALLAM-7B-Instruct & \cmark & 0.007 & 0.002 & 0.001 & 0.000 & 0.018 & 1.141 & 0.381 & 0.365 & 0.372 \\
    ALLAM-7B-Instruct & \xmark & 0.004 & 0.000 & 0.000 & 0.000 & 0.014 & 0.460 & 0.361 & 0.358 & 0.359 \\
\bottomrule
\end{tabular}
}
\caption{Statistics for the remaining evaluation metrics (ROUGE-1, ROUGE-2, ROUGE-3, ROUGE-4, METEOR, CHRF++, and BERTScore-P/R/F) on the Algerian Arabic-France (ar\_DZ-FR) language pair, comparing reasoning and non-reasoning approaches for the Text Summarization task.}
\label{tab:summary_reasoning_non_statistics_ar-dz-fr}
\renewcommand{\arraystretch}{1}
\end{table*}

\subsubsection{0-shot vs. Few-shot}
\label{sec:quantitative-results-text-summarization-shot}

The overall results for the 0-shot versus few-shot setting are reported in Table~\ref{tab:summary_shot_statistics_id-en}, Table~\ref{tab:summary_shot_statistics_jv-id-en}, Table~\ref{tab:summary_shot_statistics_su-id-en}, Table~\ref{tab:summary_shot_statistics_ha-en}, Table~\ref{tab:summary_shot_statistics_ar-dz-fr},

\begin{table*}[!th]
\centering
\resizebox{\textwidth}{!}{
\begin{tabular}{@{}lcccccccccc@{}}
\toprule
& & \multicolumn{9}{c}{\textsc{ID-EN}}\\
\cmidrule(lr){3-11}
\textbf{model} & \textbf{N-shot} & 
    \textbf{ROUGE1} &
    \textbf{ROUGE2} &
    \textbf{ROUGE3} &
    \textbf{ROUGE4} &
    \textbf{METEOR} &
    \textbf{CHRF++} &
    \textbf{BERTScore-p} &
    \textbf{BERTScore-r} &
    \textbf{BERTScore-f} \\
\midrule
\textbf{Global} \\
\midrule
    Qwen2.5-3B-Instruct & 0 & 0.343 & 0.084 & 0.029 & 0.010 & 0.263 & 36.575 & 0.896 & 0.902 & 0.894 \\
    Qwen2.5-3B-Instruct & 1 & 0.364 & 0.102 & 0.038 & 0.014 & 0.270 & 36.502 & 0.903 & 0.904 & 0.899 \\
    Qwen2.5-3B-Instruct & 3 & 0.364 & 0.102 & 0.040 & 0.014 & 0.273 & 37.287 & 0.903 & 0.905 & 0.900 \\
    Qwen2.5-7B-Instruct & 0 & 0.328 & 0.090 & 0.033 & 0.013 & 0.274 & 36.315 & 0.878 & 0.905 & 0.886 \\
    Qwen2.5-7B-Instruct & 1 & 0.368 & 0.117 & 0.047 & 0.017 & 0.272 & 36.228 & 0.906 & 0.905 & 0.901 \\
    Qwen2.5-7B-Instruct & 3 & 0.377 & 0.120 & 0.051 & 0.021 & 0.275 & 36.748 & 0.909 & 0.906 & 0.902 \\
    Qwen3-4B & 0 & 0.303 & 0.087 & 0.034 & 0.015 & 0.254 & 34.148 & 0.881 & 0.903 & 0.887 \\
    Qwen3-4B & 1 & 0.301 & 0.093 & 0.038 & 0.015 & 0.219 & 30.319 & 0.891 & 0.883 & 0.883 \\
    Qwen3-4B & 3 & 0.177 & 0.057 & 0.024 & 0.009 & 0.136 & 18.194 & 0.854 & 0.842 & 0.845 \\
    Qwen3-8B & 0 & 0.250 & 0.073 & 0.028 & 0.011 & 0.229 & 29.557 & 0.866 & 0.902 & 0.879 \\
    Qwen3-8B & 1 & 0.366 & 0.114 & 0.044 & 0.018 & 0.267 & 37.242 & 0.907 & 0.906 & 0.901 \\
    Qwen3-8B & 3 & 0.359 & 0.112 & 0.046 & 0.018 & 0.264 & 36.605 & 0.906 & 0.905 & 0.900 \\
    Aya23-8B & 0 & 0.347 & 0.098 & 0.037 & 0.015 & 0.275 & 38.158 & 0.898 & 0.904 & 0.896 \\
    Aya23-8B & 1 & 0.370 & 0.112 & 0.043 & 0.018 & 0.276 & 38.069 & 0.903 & 0.905 & 0.900 \\
    Aya23-8B & 3 & 0.374 & 0.111 & 0.047 & 0.018 & 0.280 & 38.010 & 0.904 & 0.906 & 0.900 \\
    Gemma2-9B-Instruct & 0 & 0.396 & 0.126 & 0.054 & 0.025 & 0.303 & 40.285 & 0.905 & 0.910 & 0.903 \\
    Gemma2-9B-Instruct & 1 & 0.408 & 0.139 & 0.061 & 0.028 & 0.310 & 40.324 & 0.910 & 0.911 & 0.906 \\
    Gemma2-9B-Instruct & 3 & 0.406 & 0.131 & 0.058 & 0.026 & 0.302 & 39.897 & 0.910 & 0.910 & 0.906 \\
    Gemma3-4B-Instruct & 0 & 0.287 & 0.079 & 0.030 & 0.012 & 0.270 & 36.126 & 0.847 & 0.905 & 0.872 \\
    Gemma3-4B-Instruct & 1 & 0.348 & 0.101 & 0.039 & 0.016 & 0.298 & 40.053 & 0.871 & 0.906 & 0.885 \\
    Gemma3-4B-Instruct & 3 & 0.357 & 0.105 & 0.039 & 0.015 & 0.298 & 40.241 & 0.873 & 0.906 & 0.886 \\
\midrule
\textbf{Regional} \\
\midrule
    Sailor2-8B & 0 & 0.278 & 0.067 & 0.023 & 0.007 & 0.203 & 29.451 & 0.732 & 0.746 & 0.736 \\
    Sailor2-8B & 1 & 0.240 & 0.063 & 0.023 & 0.008 & 0.180 & 25.247 & 0.636 & 0.646 & 0.639 \\
    Sailor2-8B & 3 & 0.159 & 0.043 & 0.015 & 0.004 & 0.119 & 16.885 & 0.412 & 0.419 & 0.414 \\
    Sahabat-AI-Gemma & 0 & 0.370 & 0.122 & 0.048 & 0.021 & 0.293 & 39.021 & 0.901 & 0.907 & 0.899 \\
    Sahabat-AI-Gemma & 1 & 0.389 & 0.134 & 0.061 & 0.027 & 0.292 & 37.878 & 0.912 & 0.908 & 0.906 \\
    Sahabat-AI-Gemma & 3 & 0.385 & 0.128 & 0.056 & 0.025 & 0.286 & 38.475 & 0.911 & 0.907 & 0.904 \\
\bottomrule
\end{tabular}
}
\caption{Statistics for the remaining evaluation metrics (ROUGE-1, ROUGE-2, ROUGE-3, ROUGE-4, METEOR, CHRF++, and BERTScore-P/R/F) on the Indonesian-English (ID-EN) language pair, comparing 0-shot and few-shot approaches for the Text Summarization task.}
\label{tab:summary_shot_statistics_id-en}
\end{table*}

\begin{table*}[!th]
\centering
\resizebox{\textwidth}{!}{
\begin{tabular}{@{}lcccccccccc@{}}
\toprule
& & \multicolumn{9}{c}{\textsc{JV-ID-EN}}\\
\cmidrule(lr){3-11}
\textbf{model} & \textbf{N-shot} & 
    \textbf{ROUGE1} &
    \textbf{ROUGE2} &
    \textbf{ROUGE3} &
    \textbf{ROUGE4} &
    \textbf{METEOR} &
    \textbf{CHRF++} &
    \textbf{BERTScore-p} &
    \textbf{BERTScore-r} &
    \textbf{BERTScore-f} \\
\midrule
\textbf{Global} \\
\midrule
    Qwen2.5-3B-Instruct & 0 & 0.345 & 0.084 & 0.030 & 0.012 & 0.251 & 37.283 & 0.895 & 0.894 & 0.893 \\
    Qwen2.5-3B-Instruct & 1 & 0.371 & 0.117 & 0.060 & 0.036 & 0.263 & 38.067 & 0.903 & 0.898 & 0.899 \\
    Qwen2.5-3B-Instruct & 3 & 0.372 & 0.121 & 0.060 & 0.032 & 0.273 & 38.564 & 0.904 & 0.899 & 0.901 \\
    Qwen2.5-7B-Instruct & 0 & 0.311 & 0.079 & 0.029 & 0.010 & 0.247 & 36.214 & 0.882 & 0.892 & 0.885 \\
    Qwen2.5-7B-Instruct & 1 & 0.361 & 0.108 & 0.047 & 0.022 & 0.243 & 36.253 & 0.905 & 0.896 & 0.899 \\
    Qwen2.5-7B-Instruct & 3 & 0.374 & 0.122 & 0.058 & 0.029 & 0.255 & 36.736 & 0.909 & 0.899 & 0.903 \\
    Qwen3-4B & 0 & 0.321 & 0.086 & 0.034 & 0.014 & 0.251 & 35.304 & 0.885 & 0.895 & 0.888 \\
    Qwen3-4B & 1 & 0.329 & 0.113 & 0.061 & 0.038 & 0.240 & 34.266 & 0.890 & 0.881 & 0.885 \\
    Qwen3-4B & 3 & 0.392 & 0.129 & 0.062 & 0.035 & 0.287 & 40.340 & 0.907 & 0.903 & 0.904 \\
    Qwen3-8B & 0 & 0.238 & 0.060 & 0.022 & 0.008 & 0.202 & 28.125 & 0.860 & 0.891 & 0.873 \\
    Qwen3-8B & 1 & 0.357 & 0.097 & 0.038 & 0.016 & 0.241 & 36.627 & 0.904 & 0.896 & 0.898 \\
    Qwen3-8B & 3 & 0.381 & 0.131 & 0.061 & 0.030 & 0.257 & 36.889 & 0.910 & 0.898 & 0.903 \\
    Aya23-8B & 0 & 0.349 & 0.092 & 0.034 & 0.012 & 0.252 & 38.227 & 0.897 & 0.892 & 0.894 \\
    Aya23-8B & 1 & 0.369 & 0.119 & 0.055 & 0.030 & 0.273 & 39.667 & 0.900 & 0.898 & 0.898 \\
    Aya23-8B & 3 & 0.376 & 0.123 & 0.057 & 0.031 & 0.283 & 40.235 & 0.901 & 0.900 & 0.899 \\
    Gemma2-9B-Instruct & 0 & 0.369 & 0.108 & 0.043 & 0.017 & 0.270 & 38.801 & 0.901 & 0.898 & 0.898 \\
    Gemma2-9B-Instruct & 1 & 0.387 & 0.119 & 0.047 & 0.019 & 0.276 & 39.167 & 0.906 & 0.900 & 0.902 \\
    Gemma2-9B-Instruct & 3 & 0.389 & 0.123 & 0.050 & 0.019 & 0.279 & 39.567 & 0.908 & 0.901 & 0.903 \\
    Gemma3-4B-Instruct & 0 & 0.238 & 0.059 & 0.020 & 0.007 & 0.228 & 32.902 & 0.826 & 0.895 & 0.857 \\
    Gemma3-4B-Instruct & 1 & 0.330 & 0.086 & 0.030 & 0.011 & 0.278 & 38.882 & 0.867 & 0.900 & 0.881 \\
    Gemma3-4B-Instruct & 3 & 0.331 & 0.095 & 0.039 & 0.017 & 0.279 & 39.480 & 0.867 & 0.900 & 0.882 \\
\midrule
\textbf{Regional} \\
\midrule
    Sailor2-8B & 0 & 0.225 & 0.047 & 0.015 & 0.004 & 0.156 & 25.473 & 0.658 & 0.664 & 0.660 \\
    Sailor2-8B & 1 & 0.203 & 0.052 & 0.019 & 0.006 & 0.139 & 22.138 & 0.573 & 0.576 & 0.574 \\
    Sailor2-8B & 3 & 0.135 & 0.034 & 0.013 & 0.005 & 0.091 & 14.443 & 0.368 & 0.370 & 0.369 \\
    Sahabat-AI-Gemma & 0 & 0.354 & 0.107 & 0.044 & 0.018 & 0.256 & 37.021 & 0.900 & 0.893 & 0.895 \\
    Sahabat-AI-Gemma & 1 & 0.382 & 0.133 & 0.064 & 0.035 & 0.267 & 36.991 & 0.910 & 0.897 & 0.902 \\
    Sahabat-AI-Gemma & 3 & 0.398 & 0.148 & 0.077 & 0.042 & 0.279 & 38.124 & 0.912 & 0.899 & 0.904 \\
\bottomrule
\end{tabular}
}
\caption{Statistics for the remaining evaluation metrics (ROUGE-1, ROUGE-2, ROUGE-3, ROUGE-4, METEOR, CHRF++, and BERTScore-P/R/F) on the Javanese-Indonesian-English (JV-ID-EN) language pair, comparing 0-shot and few-shot approaches for the Text Summarization task.}
\label{tab:summary_shot_statistics_jv-id-en}
\end{table*}

\begin{table*}[!th]
\centering
\resizebox{\textwidth}{!}{
\begin{tabular}{@{}lcccccccccc@{}}
\toprule
& & \multicolumn{9}{c}{\textsc{SU-ID-EN}}\\
\cmidrule(lr){3-11}
\textbf{model} & \textbf{N-shot} & 
    \textbf{ROUGE1} &
    \textbf{ROUGE2} &
    \textbf{ROUGE3} &
    \textbf{ROUGE4} &
    \textbf{METEOR} &
    \textbf{CHRF++} &
    \textbf{BERTScore-p} &
    \textbf{BERTScore-r} &
    \textbf{BERTScore-f} \\
\midrule
\textbf{Global} \\
\midrule
    Qwen2.5-3B-Instruct & 0 & 0.394 & 0.119 & 0.046 & 0.018 & 0.278 & 39.479 & 0.908 & 0.898 & 0.901 \\
    Qwen2.5-3B-Instruct & 1 & 0.372 & 0.114 & 0.044 & 0.018 & 0.257 & 36.972 & 0.910 & 0.896 & 0.901 \\
    Qwen2.5-3B-Instruct & 3 & 0.399 & 0.123 & 0.051 & 0.022 & 0.275 & 38.956 & 0.910 & 0.898 & 0.902 \\
    Qwen2.5-7B-Instruct & 0 & 0.372 & 0.121 & 0.047 & 0.019 & 0.276 & 38.705 & 0.897 & 0.898 & 0.895 \\
    Qwen2.5-7B-Instruct & 1 & 0.382 & 0.130 & 0.055 & 0.025 & 0.252 & 36.277 & 0.916 & 0.896 & 0.904 \\
    Qwen2.5-7B-Instruct & 3 & 0.408 & 0.138 & 0.057 & 0.025 & 0.273 & 38.435 & 0.916 & 0.899 & 0.906 \\
    Qwen3-4B & 0 & 0.392 & 0.133 & 0.053 & 0.024 & 0.284 & 37.895 & 0.904 & 0.900 & 0.900 \\
    Qwen3-4B & 1 & 0.270 & 0.093 & 0.043 & 0.020 & 0.192 & 26.580 & 0.881 & 0.854 & 0.866 \\
    Qwen3-4B & 3 & 0.089 & 0.028 & 0.012 & 0.005 & 0.065 & 9.357 & 0.834 & 0.795 & 0.813 \\
    Qwen3-8B & 0 & 0.331 & 0.110 & 0.047 & 0.023 & 0.256 & 33.668 & 0.892 & 0.898 & 0.892 \\
    Qwen3-8B & 1 & 0.404 & 0.132 & 0.054 & 0.024 & 0.273 & 38.908 & 0.916 & 0.900 & 0.905 \\
    Qwen3-8B & 3 & 0.420 & 0.143 & 0.060 & 0.028 & 0.285 & 39.631 & 0.919 & 0.902 & 0.908 \\
    Aya23-8B & 0 & 0.404 & 0.134 & 0.055 & 0.023 & 0.281 & 40.073 & 0.911 & 0.898 & 0.902 \\
    Aya23-8B & 1 & 0.375 & 0.114 & 0.044 & 0.019 & 0.269 & 39.839 & 0.905 & 0.896 & 0.898 \\
    Aya23-8B & 3 & 0.385 & 0.124 & 0.051 & 0.022 & 0.278 & 40.240 & 0.906 & 0.897 & 0.900 \\
    Gemma2-9B-Instruct & 0 & 0.441 & 0.161 & 0.074 & 0.037 & 0.309 & 41.759 & 0.917 & 0.904 & 0.908 \\
    Gemma2-9B-Instruct & 1 & 0.431 & 0.156 & 0.071 & 0.036 & 0.306 & 41.826 & 0.919 & 0.904 & 0.909 \\
    Gemma2-9B-Instruct & 3 & 0.453 & 0.170 & 0.080 & 0.040 & 0.319 & 42.777 & 0.920 & 0.906 & 0.910 \\
    Gemma3-4B-Instruct & 0 & 0.328 & 0.110 & 0.046 & 0.021 & 0.294 & 38.784 & 0.852 & 0.905 & 0.875 \\
    Gemma3-4B-Instruct & 1 & 0.387 & 0.128 & 0.053 & 0.026 & 0.319 & 42.294 & 0.879 & 0.906 & 0.889 \\
    Gemma3-4B-Instruct & 3 & 0.380 & 0.124 & 0.051 & 0.024 & 0.308 & 41.509 & 0.873 & 0.904 & 0.886 \\
\midrule
\textbf{Regional} \\
\midrule
    Sailor2-8B & 0 & 0.258 & 0.075 & 0.026 & 0.010 & 0.172 & 24.393 & 0.588 & 0.590 & 0.588 \\
    Sailor2-8B & 1 & 0.224 & 0.064 & 0.021 & 0.008 & 0.152 & 22.468 & 0.545 & 0.545 & 0.544 \\
    Sailor2-8B & 3 & 0.099 & 0.031 & 0.013 & 0.006 & 0.070 & 9.829 & 0.237 & 0.237 & 0.236 \\
    Sahabat-AI-Gemma & 0 & 0.448 & 0.163 & 0.073 & 0.037 & 0.309 & 41.698 & 0.917 & 0.902 & 0.907 \\
    Sahabat-AI-Gemma & 1 & 0.435 & 0.171 & 0.083 & 0.046 & 0.304 & 40.211 & 0.923 & 0.903 & 0.910 \\
    Sahabat-AI-Gemma & 3 & 0.450 & 0.177 & 0.084 & 0.042 & 0.312 & 41.001 & 0.923 & 0.903 & 0.910 \\
\bottomrule
\end{tabular}
}
\caption{Statistics for the remaining evaluation metrics (ROUGE-1, ROUGE-2, ROUGE-3, ROUGE-4, METEOR, CHRF++, and BERTScore-P/R/F) on the Sundanese-Indonesian-English (SU-ID-EN) language pair, comparing 0-shot and few-shot approaches for the Text Summarization task.}
\label{tab:summary_shot_statistics_su-id-en}
\end{table*}

\begin{table*}[!th]
\centering
\resizebox{\textwidth}{!}{
\begin{tabular}{@{}lcccccccccc@{}}
\toprule
& & \multicolumn{9}{c}{\textsc{HA-EN}}\\
\cmidrule(lr){3-11}
\textbf{model} & \textbf{N-shot} & 
    \textbf{ROUGE1} &
    \textbf{ROUGE2} &
    \textbf{ROUGE3} &
    \textbf{ROUGE4} &
    \textbf{METEOR} &
    \textbf{CHRF++} &
    \textbf{BERTScore-p} &
    \textbf{BERTScore-r} &
    \textbf{BERTScore-f} \\
\midrule
\textbf{Global} \\
\midrule
    Qwen2.5-3B-Instruct & 0 & 0.063 & 0.009 & 0.002 & 0.001 & 0.071 & 9.568 & 0.758 & 0.802 & 0.777 \\
    Qwen2.5-3B-Instruct & 1 & 0.207 & 0.041 & 0.015 & 0.005 & 0.151 & 21.180 & 0.829 & 0.848 & 0.837 \\
    Qwen2.5-3B-Instruct & 3 & 0.269 & 0.062 & 0.027 & 0.014 & 0.185 & 26.410 & 0.845 & 0.856 & 0.849 \\
    Qwen2.5-7B-Instruct & 0 & 0.176 & 0.029 & 0.009 & 0.003 & 0.131 & 18.438 & 0.814 & 0.835 & 0.822 \\
    Qwen2.5-7B-Instruct & 1 & 0.284 & 0.064 & 0.025 & 0.010 & 0.170 & 23.237 & 0.860 & 0.853 & 0.855 \\
    Qwen2.5-7B-Instruct & 3 & 0.275 & 0.063 & 0.027 & 0.012 & 0.172 & 24.478 & 0.852 & 0.852 & 0.851 \\
    Qwen3-4B & 0 & 0.033 & 0.003 & 0.001 & 0.000 & 0.040 & 6.047 & 0.740 & 0.761 & 0.748 \\
    Qwen3-4B & 1 & 0.137 & 0.026 & 0.008 & 0.003 & 0.111 & 14.864 & 0.788 & 0.805 & 0.794 \\
    Qwen3-4B & 3 & 0.139 & 0.038 & 0.018 & 0.010 & 0.110 & 15.138 & 0.785 & 0.802 & 0.791 \\
    Qwen3-8B & 0 & 0.020 & 0.002 & 0.000 & 0.000 & 0.036 & 5.450 & 0.751 & 0.762 & 0.754 \\
    Qwen3-8B & 1 & 0.060 & 0.013 & 0.005 & 0.002 & 0.075 & 8.675 & 0.756 & 0.788 & 0.769 \\
    Qwen3-8B & 3 & 0.163 & 0.034 & 0.017 & 0.008 & 0.124 & 17.889 & 0.798 & 0.824 & 0.809 \\
    Aya23-8B & 0 & 0.097 & 0.017 & 0.005 & 0.002 & 0.097 & 14.827 & 0.792 & 0.820 & 0.803 \\
    Aya23-8B & 1 & 0.082 & 0.013 & 0.003 & 0.001 & 0.093 & 11.774 & 0.775 & 0.815 & 0.792 \\
    Aya23-8B & 3 & 0.219 & 0.038 & 0.013 & 0.002 & 0.151 & 22.700 & 0.821 & 0.842 & 0.830 \\
    Gemma2-9B-Instruct & 0 & 0.360 & 0.098 & 0.036 & 0.015 & 0.243 & 33.598 & 0.884 & 0.878 & 0.879 \\
    Gemma2-9B-Instruct & 1 & 0.397 & 0.126 & 0.053 & 0.024 & 0.273 & 35.381 & 0.893 & 0.885 & 0.888 \\
    Gemma2-9B-Instruct & 3 & 0.397 & 0.128 & 0.055 & 0.026 & 0.272 & 35.009 & 0.895 & 0.885 & 0.888 \\
    Gemma3-4B-Instruct & 0 & 0.160 & 0.032 & 0.009 & 0.003 & 0.150 & 19.467 & 0.805 & 0.856 & 0.827 \\
    Gemma3-4B-Instruct & 1 & 0.196 & 0.048 & 0.017 & 0.006 & 0.175 & 21.231 & 0.806 & 0.862 & 0.830 \\
    Gemma3-4B-Instruct & 3 & 0.220 & 0.053 & 0.019 & 0.007 & 0.185 & 23.459 & 0.816 & 0.865 & 0.837 \\
\bottomrule
\end{tabular}
}
\caption{Statistics for the remaining evaluation metrics (ROUGE-1, ROUGE-2, ROUGE-3, ROUGE-4, METEOR, CHRF++, and BERTScore-P/R/F) on the Hausa-English (HA-EN) language pair, comparing 0-shot and few-shot approaches for the Text Summarization task.}
\label{tab:summary_shot_statistics_ha-en}
\end{table*}

\begin{table*}[!th]
\centering
\resizebox{\textwidth}{!}{
\begin{tabular}{@{}lcccccccccc@{}}
\toprule
& & \multicolumn{9}{c}{\textsc{ar\_DZ-FR}}\\
\cmidrule(lr){3-11}
\textbf{model} & \textbf{N-shot} & 
    \textbf{ROUGE1} &
    \textbf{ROUGE2} &
    \textbf{ROUGE3} &
    \textbf{ROUGE4} &
    \textbf{METEOR} &
    \textbf{CHRF++} &
    \textbf{BERTScore-p} &
    \textbf{BERTScore-r} &
    \textbf{BERTScore-f} \\
\midrule
\textbf{Global} \\
\midrule
    Qwen2.5-3B-Instruct & 0 & 0.006 & 0.001 & 0.001 & 0.001 & 0.032 & 1.184 & 0.807 & 0.766 & 0.784 \\
    Qwen2.5-3B-Instruct & 1 & 0.130 & 0.019 & 0.007 & 0.004 & 0.114 & 18.052 & 0.825 & 0.845 & 0.834 \\
    Qwen2.5-3B-Instruct & 3 & 0.164 & 0.019 & 0.004 & 0.001 & 0.142 & 20.782 & 0.838 & 0.855 & 0.846 \\
    Qwen2.5-7B-Instruct & 0 & 0.083 & 0.015 & 0.005 & 0.002 & 0.091 & 11.835 & 0.797 & 0.830 & 0.812 \\
    Qwen2.5-7B-Instruct & 1 & 0.153 & 0.021 & 0.007 & 0.002 & 0.132 & 21.770 & 0.837 & 0.855 & 0.845 \\
    Qwen2.5-7B-Instruct & 3 & 0.169 & 0.031 & 0.011 & 0.003 & 0.150 & 21.013 & 0.829 & 0.852 & 0.839 \\
    Qwen3-4B & 0 & 0.069 & 0.015 & 0.005 & 0.002 & 0.083 & 10.413 & 0.784 & 0.823 & 0.802 \\
    Qwen3-4B & 1 & 0.052 & 0.005 & 0.002 & 0.000 & 0.050 & 8.575 & 0.812 & 0.776 & 0.791 \\
    Qwen3-4B & 3 & 0.029 & 0.003 & 0.001 & 0.000 & 0.032 & 5.063 & 0.810 & 0.749 & 0.776 \\
    Qwen3-8B & 0 & 0.018 & 0.003 & 0.001 & 0.000 & 0.036 & 4.915 & 0.759 & 0.794 & 0.774 \\
    Qwen3-8B & 1 & 0.015 & 0.002 & 0.001 & 0.000 & 0.025 & 4.285 & 0.799 & 0.757 & 0.774 \\
    Qwen3-8B & 3 & 0.009 & 0.000 & 0.000 & 0.000 & 0.013 & 2.956 & 0.808 & 0.734 & 0.767 \\
    Aya23-8B & 0 & 0.042 & 0.010 & 0.003 & 0.001 & 0.068 & 9.450 & 0.798 & 0.779 & 0.786 \\
    Aya23-8B & 1 & 0.137 & 0.007 & 0.001 & 0.000 & 0.119 & 20.754 & 0.830 & 0.845 & 0.837 \\
    Aya23-8B & 3 & 0.142 & 0.007 & 0.001 & 0.000 & 0.126 & 20.508 & 0.831 & 0.852 & 0.840 \\
    Gemma2-9B-Instruct & 0 & 0.108 & 0.019 & 0.007 & 0.002 & 0.107 & 16.018 & 0.810 & 0.817 & 0.812 \\
    Gemma2-9B-Instruct & 1 & 0.109 & 0.017 & 0.006 & 0.002 & 0.106 & 16.043 & 0.807 & 0.835 & 0.819 \\
    Gemma2-9B-Instruct & 3 & 0.119 & 0.021 & 0.008 & 0.003 & 0.116 & 16.058 & 0.806 & 0.837 & 0.819 \\
    Gemma3-4B-Instruct & 0 & 0.020 & 0.004 & 0.002 & 0.001 & 0.032 & 2.463 & 0.758 & 0.776 & 0.764 \\
    Gemma3-4B-Instruct & 1 & 0.060 & 0.010 & 0.003 & 0.001 & 0.079 & 11.229 & 0.779 & 0.841 & 0.807 \\
    Gemma3-4B-Instruct & 3 & 0.093 & 0.014 & 0.004 & 0.001 & 0.108 & 14.957 & 0.798 & 0.846 & 0.820 \\
\midrule
\textbf{Regional} \\
\midrule
    SILMA-9B-Instruct & 0 & 0.010 & 0.002 & 0.000 & 0.000 & 0.010 & 1.230 & 0.715 & 0.756 & 0.733 \\
    SILMA-9B-Instruct & 1 & 0.041 & 0.007 & 0.002 & 0.000 & 0.027 & 4.038 & 0.740 & 0.779 & 0.756 \\
    SILMA-9B-Instruct & 3 & 0.062 & 0.008 & 0.002 & 0.001 & 0.049 & 6.571 & 0.758 & 0.788 & 0.771 \\
    ALLAM-7B-Instruct & 0 & 0.004 & 0.000 & 0.000 & 0.000 & 0.014 & 0.460 & 0.361 & 0.358 & 0.359 \\
    ALLAM-7B-Instruct & 1 & 0.011 & 0.002 & 0.001 & 0.001 & 0.014 & 1.410 & 0.204 & 0.207 & 0.205 \\
    ALLAM-7B-Instruct & 3 & 0.000 & 0.000 & 0.000 & 0.000 & 0.000 & 0.010 & 0.008 & 0.008 & 0.008 \\
\bottomrule
\end{tabular}
}
\caption{Statistics for the remaining evaluation metrics (ROUGE-1, ROUGE-2, ROUGE-3, ROUGE-4, METEOR, CHRF++, and BERTScore-P/R/F) on the Algerian Arabic-French (ar\_DZ-FR) language pair, comparing 0-shot and few-shot approaches for the Text Summarization task.}
\label{tab:summary_shot_statistics_ar-dz-fr}
\end{table*}



\subsection{Human-written vs. Machine-generated}
\label{sec:quantitative-results-human-vs-machine}

Table~\ref{tab:full_conversational_statistics} reports the overall results comparing human-written vs. machine generated conversation.

\begin{table*}[!th]
\centering
\resizebox{\textwidth}{!}{
\begin{tabular}{@{}lcccccccccc@{}}
\toprule
& \multicolumn{2}{c}{\textsc{ID-EN}} & \multicolumn{2}{c}{\textsc{JV-ID-EN}} & \multicolumn{2}{c}{\textsc{SU-ID-EN}} & \multicolumn{2}{c}{\textsc{HA-EN}} & \multicolumn{2}{c}{\textsc{ar\_DZ-FR}}\\
\cmidrule(lr){2-3} \cmidrule(lr){4-5} \cmidrule(lr){6-7} \cmidrule(lr){8-9} \cmidrule(lr){10-11}
\textbf{Metric} & 
    \textbf{Human-written} & 
    \textbf{Machine-gen} & 
    \textbf{Human-written} & 
    \textbf{Machine-gen} & 
    \textbf{Human-written} & 
    \textbf{Machine-gen} & 
    \textbf{Human-written} & 
    \textbf{Machine-gen} & 
    \textbf{Human-written} & 
    \textbf{Machine-gen} \\
\midrule
\textbf{2 speakers (50 dialogue)} \\
\midrule
    Avg length variance (tokens) & 83.436 & 29.438 & 47.58 & 43.376 & 104.202 & 27.267 & 48.252 & 28.469 & 49.231 & 47151 \\
    Total replies & 482 & 1 & 1049 & 9 & 831 & 1 & 747 & 2 & 182 & 0 \\
    Avg degree of reply distance & 3.053 & 0.02 & 2.972 & 0.18 & 2.066 & 0.02 & 2.645 & 0.02 & 2.909 & 0 \\
    Avg imbalance ratio of speaker turns & 1.364 & 1.016 & 1.315 & 1.019 & 1.368 & 1.017 & 1.195 & 1.017 & 1.587 & 1.026 \\
    Avg CMI & 0.491 & 0.711 & 0.467 & 0.733 & 0.734 & 0.719 & 0.37 & 0.374 & 0.565 & 0.169 \\
    Avg SPF & 0.306 & 0.423 & 0.312 & 0.441 & 0.461 & 0.436 & 0.202 & 0.189 & 0.251 & 0.075 \\
    Human preference & 2.721 & 2.694 & 2.627 & 2.373 & 2.840 & 2.120 & 2.483 & 2.470 & 2.969 & 1.208 \\
\midrule
\textbf{3 speakers (25 dialogue)} \\
\midrule
    Avg length variance (tokens) & 44.122 & 22.732 & 50.32 & 42.88 & 98.882 & 23.811 & 48.519 & 23.069 & 54.157 & 38.657 \\
    Total replies & 414 & 23 & 772 & 17 & 941 & 22 & 761 & 20 & 270 & 17 \\
    Avg degree of reply distance & 4.309 & 0.98 & 3.814 & 0.558 & 3.222 & 0.668 & 3.188 & 0.723 & 3.921 & 0.76 \\
    Avg imbalance ratio of speaker turns & 3.286 & 1.053 & 1.895 & 1.107 & 1.832 & 1.044 & 3.324 & 1.052 & 4.347 & 1.024 \\
    Avg CMI & 0.444 & 0.704 & 0.476 & 0.689 & 0.77 & 0.718 & 0.348 & 0.202 & 0.565 & 0.117 \\
    Avg SPF & 0.291 & 0.429 & 0.328 & 0.411 & 0.469 & 0.437 & 0.187 & 0.1 & 0.248 & 0.044 \\
    Human preference & 2.692 & 2.538 & 2.603 & 2.359 & 2.893 & 2.107 & 2.467 & 2.520 & 2.942 & 1.192 \\
\midrule
\textbf{4 speakers (25 dialogue)} \\
\midrule
    Avg length variance (tokens) & 63.764 & 32.003 & 35.918 & 32.82 & 167.263 & 24.299 & 58.806 & 15.9 & 80.132 & 40.934\\
    Total replies & 718 & 43 & 950 & 39 & 1307 & 55 & 1062 & 43 & 305 & 45\\
    Avg degree of reply distance & 4.343 & 1.263 & 4.037 & 1.099 & 3.841 & 0.998 & 3.543 & 1.005 & 5.115 & 1.097\\
    Avg imbalance ratio of speaker turns & 3.459 & 1.1 & 2.586 & 1.209 & 2.3 & 1.098 & 2.786 & 1.086 & 4.272 & 1.097\\
    Avg CMI & 0.46 & 0.667 & 0.456 & 0.729 & 0.789 & 0.719 & 0.321 & 0.191 & 0.576 & 0.108\\
    Avg SPF & 0.298 & 0.393 & 0.318 & 0.44 & 0.479 & 0.452 & 0.172 & 0.096 & 0.261 & 0.041\\
    Human preference & 2.573 & 2.533 & 2.625 & 2.083 & 2.853 & 2.200 & 2.385 & 2.538 & 2.942 & 1.192\\
\bottomrule
\end{tabular}
}
\caption{Full quantitative statistics per-language combination of human written VS machine generated conversational pattern}
\label{tab:full_conversational_statistics}
\end{table*}

\section{Additional Results}

\subsection{Taxonomy of Unanswerable QA Category}
\label{sec:quantitative-results-unanswerable-category}

For the Question Answering (QA) task, annotators for each language were instructed to create up to five unanswerable questions per instance. This design ensures that each instance is represented across the five targeted categories: Negation, Antonym, Entity-Swap, Mutual-Exclusion, and Impossible-Condition. By the end of the annotation process, three language combinations (ID–EN, JV–ID–EN, and SU–ID–EN) successfully produced up to five unanswerable questions for each instance. Table~\ref{tab:qa_unanswerable_category} reports the model performance (Acc.\%) on these three language combinations across all unanswerable categories. The results indicate that \textit{Impossible-Condition} constitutes the most challenging category, whereas \textit{Negation}, \textit{Antonym}, and \textit{Entity-Swap} are relatively easier for the models to handle.

\begin{table*}[!th]
\centering
\resizebox{\textwidth}{!}{
\begin{tabular}{@{}lccccccccccccccc@{}}
\toprule
& \multicolumn{5}{c}{\textsc{ID-EN}} & \multicolumn{5}{c}{\textsc{JV-ID-EN}} & \multicolumn{5}{c}{\textsc{SU-ID-EN}} \\
\cmidrule(lr){2-6} \cmidrule(lr){7-11} \cmidrule(lr){12-16}
\textbf{Model} &
\textbf{Neg.} & \textbf{Ant.} & \textbf{Ent.} & \textbf{Mut.} & \textbf{Imp.} &
\textbf{Neg.} & \textbf{Ant.} & \textbf{Ent.} & \textbf{Mut.} & \textbf{Imp.} &
\textbf{Neg.} & \textbf{Ant.} & \textbf{Ent.} & \textbf{Mut.} & \textbf{Imp.} \\
\midrule

\textbf{Global} & \multicolumn{15}{c}{} \\
\midrule
Qwen2.5-3B-Instruct & 46.46 & 45.45 & 36.36 & 59.59 & 76.76 & 58.58 & 46.46 & 42.42 & 67.67 & 88.88 & 90.90 & 77.77 & 77.77 & 88.88 & 95.95 \\
Qwen2.5-7B-Instruct & 48.48 & 61.61 & 40.40 & 63.63 & 87.87 & 60.60 & 51.51 & 52.52 & 65.65 & 90.90 & 80.80 & 68.68 & 76.76 & 78.78 & 91.91 \\
Qwen3-4B & 28.28 & 29.29 & 10.10 & 44.44 & 76.76 & 30.30 & 31.31 & 18.18 & 42.42 & 78.78 & 82.82 & 65.65 & 73.73 & 75.75 & 90.90 \\
Qwen3-8B & 33.83 & 44.44 & 23.23 & 73.73 & 88.88 & 50.50 & 41.41 & 33.33 & 68.68 & 90.90 & 93.93 & 86.86 & 86.86 & 85.85 & 98.98 \\
Aya23-8B & 54.54 & 44.44 & 38.38 & 64.64 & 77.77 & 50.50 & 43.43 & 38.38 & 65.65 & 81.81 & 72.72 & 66.66 & 52.52 & 73.73 & 79.79 \\
Gemma2-9B-Instruct & 35.35 & 31.31 & 20.20 & 75.75 & 87.87 & 45.45 & 36.36 & 20.20 & 74.74 & 90.90 & 92.92 & 86.86 & 87.87 & 92.92 & 94.94 \\
Gemma3-4B-Instruct & 10.10 & 6.06 & 4.04 & 18.18 & 31.31 & 13.13 & 10.10 & 7.07 & 23.23 & 34.34 & 51.51 & 26.26 & 42.42 & 40.40 & 54.54 \\
\midrule

\textbf{Regional} & \multicolumn{15}{c}{} \\
\midrule
Sailor2-8B & 10.10 & 12.12 & 3.03 & 15.15 & 16.16 & 13.13 & 7.07 & 3.03 & 16.16 & 12.12 & 41.41 & 26.26 & 23.23 & 28.28 & 34.34 \\
Sahabat-AI-Gemma & 40.40 & 25.25 & 19.19 & 74.74 & 86.86 & 44.44 & 39.39 & 18.18 & 77.77 & 91.91 & 92.92 & 85.85 & 82.82 & 91.91 & 92.92 \\

\bottomrule
\end{tabular}
}
\caption{Statistics (Acc. \%) per-language combination of Unanswerable question's category on Question Answering task. (Neg. = Negation, Ant. = Antonym, Ent. = Entity-Swap, Mut. = Mutually-Exclusion, and Imp. = Impossible-Condition)}
\label{tab:qa_unanswerable_category}
\end{table*}

\end{document}